\documentclass{article}

\usepackage[main, final]{neurips_2026}

\usepackage[utf8]{inputenc} 
\usepackage[T1]{fontenc}    
\usepackage{hyperref}       
\usepackage{url}            
\usepackage{booktabs}       
\usepackage{amsfonts}       
\usepackage{amsmath}
\usepackage{pifont}
\usepackage{nicefrac}       
\usepackage{microtype}      
\usepackage{xcolor}         
\usepackage[table]{xcolor}
\usepackage{arydshln}
\usepackage{algorithm}
\usepackage{algpseudocode}
\usepackage{graphicx}
\usepackage{cleveref}

\title{The Sparsity Whisperer}

\author{%
  \begin{minipage}[t]{0.32\textwidth}\centering
    \bfseries Linghao Kong\\
    \normalfont MIT\\
    \texttt{linghao@mit.edu}
  \end{minipage}%
  \hfill
  \begin{minipage}[t]{0.32\textwidth}\centering
    \bfseries Inimai Subramanian\\
    \normalfont MIT\\
    \texttt{inimai@mit.edu}
  \end{minipage}%
  \hfill
  \begin{minipage}[t]{0.32\textwidth}\centering
    \bfseries Micah Adler\\
    \normalfont MIT\\
    \texttt{micah432@mit.edu}
  \end{minipage}\\[4em]
  \begin{minipage}[t]{0.32\textwidth}\centering
    \bfseries Dan Alistarh\\
    \normalfont IST Austria\\
    \texttt{dan.alistarh@ist.ac.at}
  \end{minipage}%
  \hfill
  \begin{minipage}[t]{0.32\textwidth}\centering
    \bfseries Dan Gutfreund\\
    \normalfont IBM\\
    \texttt{dgutfre@us.ibm.com}
  \end{minipage}%
  \hfill
  \begin{minipage}[t]{0.32\textwidth}\centering
    \bfseries Nir Shavit\\
    \normalfont MIT \& Red Hat AI\\
    \texttt{shanir@mit.edu}
  \end{minipage}
}

\newcommand{\mymethodFOL}{Wisp}
\newcommand{\mymethodFON}{Wisp+}

\newcommand{\mymethodSO}{Whisper}

\begin{document}

\maketitle

\begin{abstract}

Pruning reduces the inference cost of large language models, but existing criteria primarily preserve large activations or reconstruct layer outputs. We argue that this overlooks a key computation performed by particularly sparsity-sensitive neurons in the MLP up and gate projections: separating similar inputs into dissimilar outputs. This suggests that effective pruning should preserve not only activations, but also the differences between outputs more broadly. We introduce a family of difference-informed pruning methods built upon this principle. \mymethodFON{} is a neuronwise, update-free method that prioritizes the input pairs each neuron separates most strongly relative to their distance. \mymethodFOL{} provides a more efficient layerwise alternative that scores weights using shared input-difference norms. Finally, \mymethodSO{} extends global difference preservation to second-order reconstruction using a lightly regularized difference Hessian. Across Llama 2 and 3.1 models from 7B to 405B parameters, our second-order variant consistently improves over strong reconstruction-based baselines, while our update-free variants improve over activation-aware baselines, especially in constrained settings. The improvements over Wanda and SparseGPT extend to structured sparsity, downstream evaluations, and other model families. Augmenting stronger techniques such as RIA and ALPS with our difference-informed criteria yields further improvements, shifting the overall accuracy-runtime frontier outward at negligible additional cost. These results suggest that preserving output differences is a broadly useful and composable signal for post-training LLM sparsification.
\makeatletter
\begingroup
  \let\@footnotemark\H@@footnotemark
  \let\@footnotetext\H@@footnotetext
  \stepcounter{footnote}%
  \protected@xdef\@thefnmark{\thefootnote}%
  \begingroup
    \renewcommand{\@makefnmark}{}
    \@footnotemark
  \endgroup
  \@footnotetext{Code available at \texttt{\href{https://github.com/Shavit-Lab/Whisper}{https://github.com/Shavit-Lab/Whisper}}.}
\endgroup
\makeatother

\end{abstract}

\section{Introduction}

As large language models (LLMs) continue to scale, their computational costs increasingly constrain deployment. This has made post-training compression an important tool for reducing inference cost while preserving the capabilities of pretrained models \citep{lin2024awq, frantar2022gptq, han2015learning, sun2024wanda, zhang2024ria, liu2023deja, lee2024cats}. Weight sparsity \citep{lecun1989optimal, hassibi1992second} is one such form of compression, removing parameters by setting selected weights to zero, which reduces the number of stored nonzero weights and offers a path to inference acceleration on hardware and kernels that exploit sparsity \citep{mishra2021accelerating, macko2025macko}. We study pruning in the one-shot setting, where a pretrained model is sparsified using only a small calibration set and no subsequent finetuning. This setting is well suited for LLMs, where the cost of retraining makes the choice of pruning criterion particularly important. 

Existing one-shot pruning methods differ primarily in how they estimate weight saliency. Magnitude pruning \citep{hagiwara1994simple}, the simplest approach, removes weights with the smallest absolute values, but degrades LLM performance sharply. Input-aware methods such as Wanda \citep{sun2024wanda}, and refinements such as RIA \citep{zhang2024ria}, scale weight values by activation statistics from a small calibration set and leave the remaining weights unchanged. SparseGPT \citep{frantar2023sparsegpt} has become a canonical second-order baseline for one-shot LLM pruning. It uses calibration inputs to form an approximate Hessian for a layerwise reconstruction problem, estimates the error induced by pruning individual weights, and updates the remaining weights to compensate. More recent methods such as ALPS \citep{meng2024alps} improve this reconstruction stage through stronger optimization. Together, these methods form a strong set of practical baselines, but they use calibration data primarily to estimate which weights support large activations or local reconstruction. This raises the question of whether other structure in the calibration set can provide useful saliency information.

A clue for such additional structure comes from Wasserstein neurons, a recently identified subpopulation of neurons that are unusually difficult to sparsify \citep{sawmya2025wasserstein, kong2025negative}. Unlike ordinary neurons, these units are characterized not primarily by large activations or large weight magnitudes, but by their input-output geometry: they tend to map similar input vectors to dissimilar output scalars. Prior work shows that this behavior is strongly associated with the Wasserstein distance (WD) of a neuron's output distribution to a Gaussian baseline, thus terming high-WD units as Wasserstein neurons. These neurons' sensitivity to pruning is especially apparent at higher sparsities and in more constrained settings. This suggests that pruning criteria may benefit from accounting for the geometry of the differences between output values more generally, rather than relying only on activation magnitude or reconstruction error.


Motivated by this insight, we develop a family of one-shot pruning methods that use output differences as an explicit saliency signal in the MLP up and gate projections, where Wasserstein neurons concentrate. The neuronwise variant of \textbf{W}asserstein-\textbf{I}nspired \textbf{S}aliency \textbf{P}runing (\mymethodFON{}) replaces Wanda's activation-magnitude statistic with a saliency based on the pairs of calibration inputs that each neuron separates most strongly relative to their distance, leaving the surviving weights unchanged. Its more efficient layerwise counterpart, \mymethodFOL{}, instead uses a shared statistic based on pairwise input differences across the layer. The second-order method, \textbf{W}asserstein-inspired \textbf{H}essian-\textbf{I}nformed \textbf{S}eparation-\textbf{P}reserving \textbf{E}rror \textbf{R}econstruction (\mymethodSO{}), retains SparseGPT's weight updates but replaces its reconstruction geometry in the same projections with a Hessian constructed from pairwise input differences to preserve differences in output. Attention and down projections retain the corresponding baseline method, so our methods alter only the relevant criteria without changing model architecture or introducing additional optimization. We also use \mymethodFOL{}, \mymethodFON{} and \mymethodSO{} to denote their underlying difference-informed saliency statistics and Hessian geometry when incorporated into other pruning methods: while introduced through Wanda and SparseGPT, these components also improve the stronger RIA and ALPS formulations.

\begin{figure}[h]
    \centering 
    \includegraphics[width=1\linewidth]{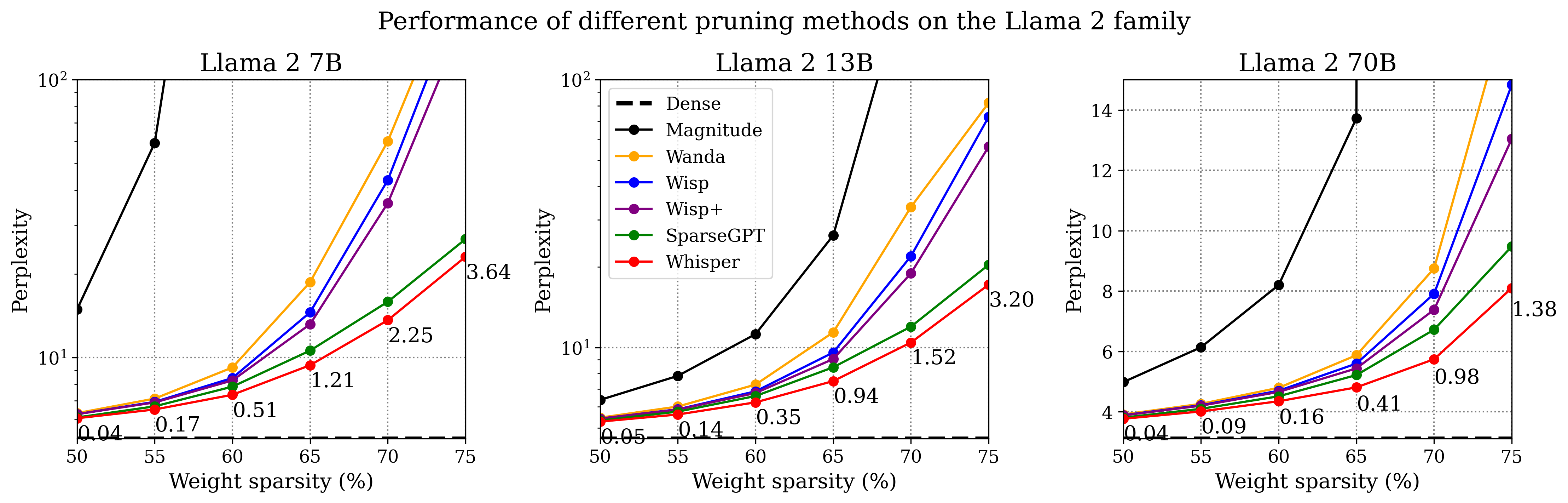}
    \caption{Unstructured sparsity scaling on Llama 2. Perplexity is shown for Llama 2 7B, 13B, and 70B as unstructured sparsity increases from 50\% to 75\% (lower is better). \mymethodFOL{} and \mymethodFON{} consistently improve over Wanda in the no-update setting, while \mymethodSO{} improves over SparseGPT at every sparsity and model size. The gains grow as pruning becomes more aggressive, suggesting that preserving pairwise output differences becomes increasingly important at high sparsity. Numbers written next to values indicate absolute perplexity improvement of \mymethodSO{} over SparseGPT.}
    \label{fig:llama_2_curve}
\end{figure}

Empirically, we find that preserving output differences improves one-shot pruning across sparsity regimes, model scales, model families, and pruning formulations. On unstructured sparsity sweeps over the Llama 2 family, \mymethodFOL{} and \mymethodFON{} consistently improve over Wanda and close the gap to SparseGPT, while \mymethodSO{} improves over SparseGPT at every tested sparsity from 50\% to 75\%, with the largest gains appearing at higher sparsities (\Cref{fig:llama_2_curve}). The same pattern holds under 2:4 structured sparsity and on downstream Open LLM Leaderboard v1-like evaluations: \mymethodFON{} is the strongest no-update method, and \mymethodSO{} obtains the best overall perplexity (PPL) and mean downstream score across the Llama 2 and 3.1 families (\Cref{tab:llama_main}). The gains generalize to Mistral and Qwen models (\Cref{tab:mistral_qwen}), and they do not come from hidden runtime costs: \mymethodFOL{} runs at Wanda speed, \mymethodFON{} remains faster than SparseGPT, and \mymethodSO{} runs at approximately SparseGPT speed (\Cref{tab:runtime}). Moreover, the difference-informed signal composes with stronger pruning techniques: augmenting RIA with Wisp or Wisp+ improves RIA across all tested settings, while augmenting ALPS with Whisper improves ALPS in all but one setting, with negligible additional pruning time (\Cref{sec:composability}). Together, these results show that output differentiation improves both update-free saliency pruning and second-order reconstruction, from canonical baselines to stronger recent methods. This consistency distinguishes the signal from method-specific saliency or solver refinements and supports separation preservation as a general principle for one-shot LLM sparsification.

\section{Related Work}

\paragraph{Post-training compression}
The high cost of LLM inference has motivated post-training compression methods such as quantization \citep{lin2024awq, ashkboos2024quarot, frantar2022gptq}, activation sparsity \citep{lee2024cats, liu2023deja}, and weight sparsity. We focus on one-shot weight pruning, where a pretrained model is sparsified using a small calibration set and no subsequent finetuning. This setting avoids retraining cost, but makes the pruning criterion especially important.

\paragraph{Canonical one-shot pruning backbones}
SparseGPT \citep{frantar2023sparsegpt}, building on the Optimal Brain Surgeon/Compression framework \citep{hassibi1992second, frantar2022optimal}, formulates pruning as a layerwise sparse reconstruction problem. Given calibration inputs, it forms an approximate Hessian $\mathbf{H}=\mathbf{X}\mathbf{X}^{T}$, prunes weights according to estimated reconstruction cost, and updates the remaining weights to compensate for pruning error. Wanda \citep{sun2024wanda} instead provides a simpler first-order alternative, scoring weights by $|\mathbf{W}_{ij}| \cdot \|\mathbf{X}_{j}\|_2$ and pruning without weight updates. These methods define two canonical practical regimes for one-shot LLM pruning: Wanda-style update-free saliency pruning and SparseGPT-style second-order reconstruction with weight updates. Our methods intervene directly in the calibration geometry of these regimes: \mymethodFOL{} and \mymethodFON{} replace Wanda's activation statistic with difference-informed saliencies, while \mymethodSO{} replaces SparseGPT's standard Hessian with a difference-informed second moment.

\paragraph{Stronger pruning formulations}
Recent methods strengthen Wanda- and SparseGPT-style pruning while remaining within comparable pruning paradigms. RIA \citep{zhang2024ria} refines update-free saliency through normalized weight importance and activation statistics, while ALPS \citep{meng2024alps} improves sparse reconstruction through a stronger but costlier optimization procedure and post-processing. We therefore use RIA and ALPS as stronger recent baselines and test composability: RIA-\mymethodFOL{} and RIA-\mymethodFON{} replace RIA's activation statistic with difference-informed statistics, while ALPS-\mymethodSO{} replaces ALPS's reconstruction Hessian with our difference-informed geometry. Other approaches, including gradient-based scoring \citep{das2023gblm}, pruning-and-growing mask refinement \citep{zhang2024dsnot}, and symbolic saliency search \citep{dong2024prunerzero} are more orthogonal to our intervention. To the best of our knowledge, prior work has not studied replacing activation-based reconstruction geometry with one that explicitly preserves pairwise output differences.

\paragraph{Wasserstein neurons and interpretability-guided sparsity}
Our work builds upon the findings of \cite{sawmya2025wasserstein} on Wasserstein neurons. They are a subpopulation of transformer MLP neurons concentrated within the up and gate projections whose output distributions have large WD from a Gaussian baseline and that map similar inputs to dissimilar outputs, a behavior quantified by their mapping difficulty metric, explained in more detail in \Cref{sec:mapping_diff}. These high-WD neurons are disproportionately sensitive to weight sparsification, with fragility increasing at higher sparsity. Prior work proposed Sparse Expansion to mitigate this high-sparsity fragility by routing inputs to multiple sparse experts, but doing so expands the sparse weights across 16 experts, making it better suited as a analytical framework than as a practical pruning method. In contrast, our methods add no additional components, instead converting output differentiation into practical pruning criteria.

\section{Methodology}

We first introduce \mymethodFON{}, which most directly captures the neuron-specific input-differentiating behavior that motivated our approach. We then develop \mymethodFOL{}, a more efficient layerwise formulation, and \mymethodSO{}, which extends global difference preservation to second-order reconstruction.

For all three methods, we apply difference-informed criteria only to the MLP up and gate projections, where Wasserstein neurons are concentrated, retaining the corresponding baseline methods for attention and down projections. This choice is supported by our component-wise ablations, which show that the benefits of difference-informed pruning are concentrated in these projections (\Cref{tab:component_analysis}).

\subsection{\mymethodFON{}: Neuronwise difference-informed saliency}

Recent work suggests that individual Wasserstein neurons specialize in differentiating distinct subsets of input pairs \citep{kong2025negative}. We therefore begin by constructing a neuron-specific saliency that prioritizes the input pairs each neuron separates most strongly relative to their distance.

Consider a linear layer $\mathbf{W}\in\mathbb{R}^{n\times m}$ with $n$ output neurons and $m$ input channels. Let $\mathbf{X}=[\mathbf{x}_1,\ldots,\mathbf{x}_s]\in\mathbb{R}^{m\times s}$
denote the calibration inputs to the layer, where each column $\mathbf{x}_\ell\in\mathbb{R}^m$ is one input vector. We write $\mathbf{X}_{j}\in\mathbb{R}^{s}$ for the $j$-th input channel across the $s$ calibration samples.

Let $\pi$ be a random permutation of $\{1,\ldots,s\}$ and let $\mathbf{X}_{\pi}=[\mathbf{x}_{\pi(1)},\ldots,\mathbf{x}_{\pi(s)}]$ be the corresponding column-permuted calibration matrix. We define $\Delta\mathbf{X}=\mathbf{X}-\mathbf{X}_{\pi}$,
so that the $\ell$-th column is the sampled pairwise difference $\Delta\mathbf{X}_{:,\ell}=\mathbf{x}_{\ell}-\mathbf{x}_{\pi(\ell)}$. More details on sampling choices in \Cref{sec:sampling}. For neuron $i$, let $y_{i,\ell}=\mathbf{w}_i\mathbf{x}_{\ell}$
denote its scalar output on calibration input $\mathbf{x}_{\ell}$. The output difference induced by the $\ell$-th sampled pair is $
\Delta y_{i,\ell}
=
y_{i,\ell}-y_{i,\pi(\ell)}
=
\mathbf{w}_i\Delta\mathbf{X}_{:,\ell}.
$
We score each sampled pair by the output separation induced per unit input distance:
$
r_{i,\ell}
=
\frac{
|\Delta y_{i,\ell}|
}{
\|\Delta\mathbf{X}_{:,\ell}\|_2
}
$. We then select the top $K$ pair indices for neuron $i$: $
\mathcal{P}_i
=
\operatorname{TopK}_{\ell\in\{1,\ldots,s\}} r_{i,\ell}
$. In all experiments in the main text, $K = 0.005s$.

Given this neuron-specific pair set, we estimate which input channels are most involved in the separations that neuron $i$ performs. For input channel $j$, we compute the average absolute channel difference over the selected pairs:
$
\frac{1}{|\mathcal{P}_i|}
\sum_{\ell\in\mathcal{P}_i}
|\Delta\mathbf{X}_{j,\ell}|
$. We use an $\ell_1$ statistic over the selected difference vectors as it performs better empirically than the corresponding $\ell_2$ statistic in the neuronwise setting. The neuronwise difference-informed saliency is then $
\mathbf{S}_{ij}^{\text{\mymethodFON}}
=
|\mathbf{W}_{ij}| \cdot \frac{1}{|\mathcal{P}_i|}
\sum_{\ell\in\mathcal{P}_i}
|\Delta\mathbf{X}_{j,\ell}|
$.

Thus, \mymethodFON{} prioritizes weights associated with the specific input pairs that each neuron maps unusually far apart relative to their distance. We further validate this connection to the motivating behavior by showing that \mymethodFON{} assigns greater saliency to weights in Wasserstein neurons and better preserves their mapping difficulty under pruning (\Cref{fig:wisp_analysis}c,d). Additional ablations for the choice of $K$ and neuron specialization proportion are provided in \Cref{sec:wisp_to_wispplus}.

\mymethodFON{} is a first-order, no-update method. While it has additional cost arising from selecting neuron-specific pairs and computing the corresponding saliencies, it remains faster than SparseGPT in both asymptotic complexity and total pruning time (\Cref{tab:comparison,tab:runtime}).

\subsection{\mymethodFOL{}: Layerwise difference-informed saliency}
\label{sec:wisp}

While \mymethodFON{} directly targets neuron-specific input differentiation, preserving output differences more generally can also provide a useful pruning signal. We therefore introduce \mymethodFOL{}, a simpler and more efficient layerwise alternative that avoids neuron-specific pair selection and does not update weights.

For a pair of inputs $\mathbf{x}_1$ and $\mathbf{x}_2$, their output difference under a linear layer is $\mathbf{W}\mathbf{x}_1-\mathbf{W}\mathbf{x}_2=\mathbf{W}(\mathbf{x}_1-\mathbf{x}_2)$. Thus, preserving pairwise output differences after pruning amounts to
preserving the layer outputs induced by differences between inputs.

Using the same calibration inputs and sampled differences $\Delta\mathbf{X}$ defined above, the output difference induced for neuron $i$ by the $\ell$-th pair is
$
\Delta y_{i,\ell}
=
\mathbf{w}_i\Delta\mathbf{X}_{:,\ell}
=
\sum_{j=1}^{m}\mathbf{W}_{ij}\Delta\mathbf{X}_{j,\ell}.
$ Thus, the contribution of weight $\mathbf{W}_{ij}$ to pairwise output separation across sampled pairs has magnitude proportional to $|\mathbf{W}_{ij}|\cdot\|\Delta\mathbf{X}_{j}\|_2$. We therefore define the layerwise difference-informed saliency
$
\mathbf{S}_{ij}^{\text{\mymethodFOL}}
=
|\mathbf{W}_{ij}|\cdot\|\Delta\mathbf{X}_{j}\|_2
$. More formally, this score also follows from a difference-preservation objective
in which pruning is chosen to preserve
$\mathbf{W}\Delta\mathbf{X}$ rather than the ordinary layer outputs
$\mathbf{W}\mathbf{X}$ (\Cref{sec:difference_derivation_wisp}).
Under a diagonal approximation and without updates to the surviving
weights, pruning $\mathbf{W}_{ij}$ incurs cost
$\mathbf{W}_{ij}^2\|\Delta\mathbf{X}_{j}\|_2^2$.
Ranking by the square root of this cost yields the saliency above.

Interestingly, the pairwise difference statistics underlying \mymethodFOL{} capture centered activation variation in expectation, up to a constant factor, yielding a deterministic implementation with nearly identical performance (\Cref{sec:centering}). We retain the pairwise framework because it directly expresses output-difference preservation and enables \mymethodFON{}'s neuron-specific pair selection, which cannot be captured by a single, shared global statistic.

To connect general difference preservation more directly to the motivating behavior of separating similar inputs, we also evaluate nearest-neighbor pairing for \mymethodFOL{}. This improves performance particularly under constrained sparsity, while \mymethodFON{} generally improves further through neuron-specific pair selection (\Cref{fig:wisp_analysis}b). These more targeted approaches incur additional computational cost, so we retain \mymethodFOL{} with random pairing as the efficient layerwise method (\Cref{sec:sampling}).

\subsection{\mymethodSO{}: Difference-informed Hessian reconstruction}
The layerwise difference statistic $\Delta \mathbf{X}$ also defines a natural second-order pruning objective. SparseGPT seeks to preserve the layer output $\mathbf{W}\mathbf{X}$ on calibration inputs, yielding the standard Hessian
$
\mathbf{H}=\mathbf{X}\mathbf{X}^{T}
$.
By contrast, our goal is to preserve the layer's pairwise output separations,
$
\mathbf{W}\Delta\mathbf{X}
$,
where $\Delta\mathbf{X}$ contains sampled differences between calibration inputs, constructed as before. Replacing the standard reconstruction target $\mathbf{W}\mathbf{X}$ with the separation-preservation target $\mathbf{W}\Delta\mathbf{X}$ yields the difference Hessian
$
\mathbf{H}_{\Delta}
=
\Delta\mathbf{X}\Delta\mathbf{X}^{T}
$.
Thus, while \mymethodFOL{} can be viewed as a first-order diagonal approximation to preserving $\mathbf{W}\Delta\mathbf{X}$, the full second-order analogue is obtained by using $\mathbf{H}_{\Delta}$ in place of $\mathbf{H}$. We provide the derivation in \Cref{sec:difference_derivation_whisper}. Similarly, this difference Hessian is equivalent up to scale to a centered covariance formulation, yielding a deterministic implementation of \mymethodSO{} that closely matches the performance of our sampled formulation (\Cref{sec:centering}).

In practice, using only $\mathbf{H}_{\Delta}$ can be unstable, as reconstruction overemphasizes preserving relative pairwise differences while discarding information about the absolute calibration output values. We therefore lightly regularize the difference Hessian with the standard activation Hessian. Let
$
\mu=\frac{\operatorname{tr}(\mathbf{H})}{m}$ and $
\mu_{\Delta}=\frac{\operatorname{tr}(\mathbf{H}_{\Delta})}{m}$. 
We first rescale $\mathbf{H}_{\Delta}$ to match the average diagonal scale of $\mathbf{H}$, and then define
$
\widetilde{\mathbf{H}}
=
\gamma \mathbf{H}
+
(1-\gamma)
\frac{\mu}{\mu_{\Delta}}
\mathbf{H}_{\Delta}
$
where $\gamma$ is small, equal to $0.01$ in all experiments. This interpolation preserves the difference-informed geometry while retaining a small amount of the standard reconstruction objective. More details about this regularization can be found in \Cref{sec:gamma_justification}.

\mymethodSO{} uses $\widetilde{\mathbf{H}}$ as a drop-in replacement for the SparseGPT Hessian. The pruning score becomes
$
\mathbf{S}_{ij}^{\text{\mymethodSO}}
=
{|\mathbf{W}_{ij}|^2}/
{\left[(\widetilde{\mathbf{H}}+\lambda\mathbf{I})^{-1}\right]_{jj}}
$
and the remaining weights are updated using the same reconstruction procedure as SparseGPT. $\lambda=0.01$ across all experiments for SparseGPT and \mymethodSO{}.

We provide the pseudocode for our methods in \Cref{sec:pseudocode} and summarize the pruning metrics and complexities of our and prior methods below. Note that for \mymethodFOL{} and \mymethodFON{}, we follow the precedent established by Wanda and prune each neuron to the same sparsity level, as we do for Wanda itself.

\begin{table}[h]
\centering
\begin{tabular}{lcccc}
\toprule
Method & Weight Update & Calibration & Pruning Metric $\mathbf{S}_{ij}$ & Complexity \\
\midrule
Magnitude & \ding{55} & \ding{55} 
& $|\mathbf{W}_{ij}|$ 
& $O(1)$ \\
\hdashline
Wanda & \ding{55} & \ding{51} 
& $|\mathbf{W}_{ij}| \cdot \|\mathbf{X}_{j}\|_2$ 
& $O(d^2_{\mathrm{hidden}})$ \\
\rowcolor{gray!20}
\mymethodFOL & \ding{55} & \ding{51} 
& $|\mathbf{W}_{ij}| \cdot \|\Delta \mathbf{X}_{j}\|_2$
& $O(d^2_{\mathrm{hidden}})$ \\
\rowcolor{gray!20}
\mymethodFON & \ding{55} & \ding{51} 
& $|\mathbf{W}_{ij}| \cdot 
\frac{1}{|\mathcal{P}_i|}
\sum_{\ell\in\mathcal{P}_i}
|\Delta \mathbf{X}_{j,\ell}|$
& $O(d^2_{\mathrm{hidden}})$ \\
\hdashline
SparseGPT & \ding{51} & \ding{51} 
& $\displaystyle {|\mathbf{W}_{ij}|^2}/
{\left[(\mathbf{H}+\lambda\mathbf{I})^{-1}\right]_{jj}}$
& $O(d^3_{\mathrm{hidden}})$ \\
\rowcolor{gray!20}
\mymethodSO & \ding{51} & \ding{51} 
& $\displaystyle {|\mathbf{W}_{ij}|^2}/
{\left[(\mathbf{\widetilde{\mathbf{H}}}+\lambda\mathbf{I})^{-1}\right]_{jj}}$
& $O(d^3_{\mathrm{hidden}})$ \\
\midrule
\end{tabular}
\caption{Comparison of difference-informed pruning methods with existing LLM pruning algorithms.}
\label{tab:comparison}
\end{table}

\subsection{Evaluation protocol}
We evaluate on Llama 2 (7B, 13B, 70B) \citep{touvron2023llama2}, Llama 3.1 (8B, 70B, 405B) \citep{dubey2024llama}, Mistral 7B v0.3 \citep{jiang2023mistral}, Qwen3 (8B, 14B) \citep{yang2025qwen3}, and Granite 4.1 (8B, 30B) \citep{granite2026}. For all methods, we use 128 sequences of 4096 tokens from the WikiText-2 \citep{merity2016pointer} training set for calibration. \mymethodFOL{}, \mymethodFON{}, and \mymethodSO{} form pairs within each calibration sequence rather than across the full set for memory efficiency, and per-sequence saliency is aggregated. We prune each linear layer to the same sparsity, skipping pruning the embedding layer and final language modeling head as is standard practice. We test both unstructured and 2:4 sparsity \citep{mishra2021accelerating}, where two of every four consecutive weights are pruned for hardware acceleration. We use NVIDIA H100 and L40S GPUs for our experiments. 

For language modeling, we report perplexity on the WikiText-2 test set. For downstream performance, we use the Language Model Evaluation Harness \citep{eval-harness} to evaluate the sparsified models on an Open LLM Leaderboard v1-like task suite: ARC Challenge (25-shot) \citep{clark2018think}, MathQA (0-shot) \citep{amini2019mathqa}, HellaSwag (10-shot) \citep{zellers2019hellaswag}, MMLU (5-shot) \citep{hendrycks2020measuring}, TruthfulQA-MC (0-shot) \citep{lin2022truthfulqa}, and WinoGrande (5-shot) \citep{sakaguchi2021winogrande}. In all tables, lower perplexity and higher benchmark score indicate better performance.

\section{Experimental Results}

\subsection{Language modeling capability scaling under unstructured sparsity}

\Cref{fig:llama_2_curve} shows perplexity across the Llama 2 family as sparsity increases from 50\% to 75\%, with raw values in \Cref{tab:sparsity_sweep}. For all three models, the same ordering emerges. \mymethodSO{} consistently performs the best overall, improving over SparseGPT. Among methods without weight updates, \mymethodFOL{} and \mymethodFON{} consistently outperform Wanda, with \mymethodFON{} usually yielding the stronger performance.

The gains are more modest at lower sparsity, where all input-aware methods remain close, but grow as pruning becomes more aggressive. In particular, \mymethodFON{} meaningfully reduces the performance gap from the baselines with and without weight updates, quantified as 
$\frac{
\mathrm{PPL}_{\mathrm{Wanda}} - \mathrm{PPL}_{\mymethodFON}
}{
\mathrm{PPL}_{\mathrm{Wanda}} - \mathrm{PPL}_{\mathrm{SparseGPT}}
}$.
At 65\% sparsity, \mymethodFON{} closes 68\% of the Wanda-SparseGPT gap on Llama 2 7B, 78\% on Llama 2 13B, and 67\% on Llama 2 70B. Thus, although \mymethodFON{} performs no weight updates, it recovers a large fraction of the benefit normally obtained by moving from first-order pruning to second-order reconstruction.

The difference-informed second-order method yields the strongest overall performance, with \mymethodSO{} improving over SparseGPT at every reported sparsity and model size. These improvements are obtained using the same SparseGPT mask selection and weight-update rule, changing only the Hessian geometry in the MLP up and gate projections. Overall, as the number of remaining weights decreases, it becomes increasingly important to preserve directions along which outputs are separated.

\subsection{Scaling across the Llama families}

We measure the WikiText-2 perplexity and downstream benchmark scores for the Llama 2 and 3.1 families at 2:4 sparsity, 50\% unstructured sparsity, and 65\% unstructured sparsity (\Cref{tab:llama_main}). The same pattern once again appears across both model families and across all sparsity settings. Among methods with weight updates, \mymethodSO{} consistently improves over SparseGPT, yielding the best perplexity score in all reported configurations and the best benchmark score in all but one, in which \mymethodFOL{} and \mymethodFON{} outperform it. Among methods without weight updates, \mymethodFOL{} improves over Wanda in nearly every setting, and \mymethodFON{} usually improves further.

\begin{table}[h]
\centering
\resizebox{\linewidth}{!}{
\begin{tabular}{lcccccccccccc}
\toprule
& & \multicolumn{6}{c}{Llama 2} & \multicolumn{5}{c}{Llama 3.1} \\
& & \multicolumn{2}{c}{7B} & \multicolumn{2}{c}{13B} & \multicolumn{2}{c}{70B} & \multicolumn{2}{c}{8B} & \multicolumn{2}{c}{70B} & \multicolumn{1}{c}{405B}\\
\midrule
Method & Sparsity & PPL & Mean & PPL & Mean & PPL & Mean & PPL & Mean & PPL & Mean & PPL \\

\midrule
Dense && $5.12$ & $53.22$ & $4.57$ & $57.01$ & $3.12$ & $65.17$ & $5.84$ & $61.52$ & $2.64$ & $70.62$ & $1.34$ \\
\midrule
Magnitude && $54.39$ & $38.16$ & $8.32$ & $43.93$ & $6.33$ & $54.83$ & $765.3$ & $32.77$ & $1.1e3$ & $39.34$ & $10.55$ \\
Wanda && $10.53$ & $39.39$ & $7.79$ & $43.94$ & $4.97$ & $57.13$ & $18.87$ & $36.81$ & $7.84$ & $56.76$ & $4.60$ \\
\mymethodFOL{} &2:4& $9.52$ & $40.41$ & $7.36$ & $45.49$ & $4.86$ & $57.57$ & $17.65$ & $38.03$ & $7.61$ & $57.82$ & $4.53$ \\
\mymethodFON{} && $\underline{9.20}$ & $\underline{40.79}$ & $\underline{7.24}$ & $\underline{46.00}$ & $\underline{4.81}$ & $\underline{57.79}$ & $\underline{16.70}$ & $\underline{38.40}$ & $\underline{7.51}$ & $\underline{58.01}$ & $\underline{4.50}$ \\
SparseGPT && $8.18$ & $41.19$ & $6.69$ & $44.93$ & $4.71$ & $57.05$ & $11.19$ & $41.62$ & $6.78$ & $56.98$ & $4.20$\\
\mymethodSO{} && $\mathbf{7.56}$ & $\mathbf{41.98}$ & $\mathbf{6.29}$ & $\mathbf{46.60}$ & $\mathbf{4.51}$ & $\mathbf{58.39}$ & $\mathbf{10.18}$ & $\mathbf{42.24}$ & $\mathbf{6.40}$ & $\mathbf{58.22}$ & $\mathbf{4.06}$\\
\midrule

Magnitude && $14.89$ & $45.25$ & $6.37$ & $51.12$ & $4.99$ & $60.54$ & $150.9$ & $38.66$ & $17.91$ & $54.14$ & $40.27$\\
Wanda && $6.31$ & $47.41$ & $5.47$ & $52.87$ & $3.91$ & $62.52$ & $8.76$ & $50.79$ & $5.29$ & $\underline{64.97}$ & $2.75$ \\
\mymethodFOL{} &50\%& $\underline{6.25}$ & $47.65$ & $\underline{5.42}$ & $\mathbf{\underline{53.28}}$ & $3.89$ & $\underline{62.81}$ & $8.62$ & $\underline{51.31}$ & $5.25$ & $64.92$ & $2.73$\\
\mymethodFON{} && $\underline{6.25}$ & $\underline{47.91}$ & $\underline{5.42}$ & $53.21$ & $\underline{3.88}$ & $62.76$ & $\underline{8.50}$ & $51.25$ & $\underline{5.23}$ & $64.95$ & $\underline{2.72}$ \\
SparseGPT && $6.09$ & $48.35$ & $5.34$ & $52.84$ & $3.81$ & $62.66$ & $7.80$ & $53.16$ & $4.94$ & $65.77$ & $2.59$ \\
\mymethodSO{} && $\mathbf{6.04}$ & $\mathbf{48.62}$ & $\mathbf{5.28}$ & $53.14$ & $\mathbf{3.77}$ & $\mathbf{62.96}$ & $\mathbf{7.64}$ & $\mathbf{53.49}$ & $\mathbf{4.93}$ & $\mathbf{65.97}$ & $\mathbf{2.57}$ \\

\midrule
Magnitude && $3.7e4$ & $32.76$ & $26.24$ & $35.87$ & $13.73$ & $47.02$ & $3.4e4$ & $\underline{33.19}$ & $8.3e3$ & $32.60$ & $1.1e4$ \\
Wanda && $18.81$ & $33.92$ & $11.33$ & $36.72$ & $5.89$ & $54.31$ & $36.48$ & $32.44$ & $9.83$ & $49.41$ & $5.64$ \\
\mymethodFOL{} &65\%& $14.61$ & $35.19$ & $9.59$ & $38.90$ & $5.60$ & $55.64$ & $34.90$ & $32.89$ & $9.25$ & $51.17$ & $5.51$ \\
\mymethodFON{} && $\underline{13.17}$ & $\underline{36.05}$ & $\underline{9.04}$ & $\underline{39.71}$ & $\underline{5.44}$ & $\underline{55.83}$ & $\underline{30.09}$ & $33.10$ & $\underline{9.01}$ & $\underline{51.55}$ & $\underline{5.48}$ \\
SparseGPT && $10.64$ & $38.36$ & $8.41$ & $41.11$ & $5.22$ & $55.76$ & $14.51$ & $38.95$ & $7.29$ & $55.34$ & $4.86$ \\
\mymethodSO{} && $\mathbf{9.46}$ & $\mathbf{39.67}$ & $\mathbf{7.54}$ & $\mathbf{42.57}$ & $\mathbf{4.83}$ & $\mathbf{57.71}$ & $\mathbf{13.07}$ & $\mathbf{39.37}$ & $\mathbf{6.88}$ & $\mathbf{56.00}$ & $\mathbf{4.68}$\\

\midrule
\end{tabular}
}
\caption{Performance on Llama 2 and 3.1. Bold indicates best overall performance and underline indicates best performance without weight updates. We report only perplexity for Llama 3.1 405B due to the prohibitive cost of downstream evaluation at this scale. Results collected over three trials.}
\label{tab:llama_main}
\end{table}

At 2:4 sparsity, \mymethodFON{} is the best no-update method for all models for both language modeling and downstream performance, indicating that the difference-informed statistic provides a stronger saliency signal even under structured settings. Overall, the reconstruction-based \mymethodSO{} yields the best perplexity and downstream performance for every model. 

The gains become more pronounced at higher sparsity. At 65\% unstructured sparsity, \mymethodFON{} substantially improves over Wanda across all models, while \mymethodSO{} improves over SparseGPT in both perplexity and mean downstream score. These results are consistent with the motivating hypothesis: as sparsity increases, preserving output-separation structure becomes increasingly important. At 50\% sparsity, the models are less stressed, but the same ordering mostly persists. \mymethodSO{} still obtains the lowest perplexity for every model, and its downstream mean is best or near-best across the table. This suggests that the difference-informed Hessian is not only useful in extreme pruning regimes, but also improves more practically relevant one-shot pruning settings. Raw scores provided in \Cref{sec:raw_scores}.

\subsection{Task-level behavior on Llama 2 70B}

\begin{table}[h]
\centering
\resizebox{\linewidth}{!}{
\begin{tabular}{lccccccccc}
\toprule
Method & Sparsity & PPL & Mean & ARC-C & MathQA & HS & MMLU & TQA & WinoG \\
\midrule
Dense && $3.12$ & $65.17$ & $67.32$ & $38.32$ & $87.27$ & $69.65$ & $44.82$ & $83.66$ \\
\midrule
Magnitude && $6.33$ & $54.83$ & $55.97$ & $30.35$ & $78.36$ & $50.73$ & $\mathbf{\underline{39.90}}$ & $73.64$ \\
Wanda && $4.97$ & $57.13$ & $\underline{59.50}$ & $31.89$ & $78.63$ & $55.96$ & $38.73$ & $78.08$ \\
\mymethodFOL{} & 2:4 & $4.86$ & $57.57$ & $59.16$ & $31.98$ & $79.41$ & $56.92$ & $39.63$ & $78.32$ \\
\mymethodFON{} && $\underline{4.81}$ & $\underline{57.79}$ & $59.04$ & $\underline{32.19}$ & $\underline{79.72}$ & $\underline{57.45}$ & $39.77$ & $\underline{78.56}$ \\
SparseGPT && $4.71$ & $57.05$ & $58.96$ & $\mathbf{32.51}$ & $78.20$ & $57.26$ & $37.43$ & $77.95$ \\
\mymethodSO{} && $\mathbf{4.51}$ & $\mathbf{58.39}$ & $\mathbf{60.69}$ & $32.45$ & $\mathbf{79.91}$ & $\mathbf{58.82}$ & $39.65$ & $\mathbf{78.80}$ \\
\midrule
\end{tabular}
}
\caption{Performance on Llama 2 70B at 2:4 sparsity. Bold indicates best performance and underline indicates best performance without weight updates. Results collected over three trials.}
\label{tab:llama_2_70B}
\end{table}

We provide a detailed task breakdown for Llama 2 70B. The improvements in \Cref{tab:llama_main} arise from multiple benchmarks rather than being driven by a single task.

\subsection{Localizing and preserving input differentiation}

Having established the benefits of difference-informed pruning, we investigate how increasingly targeted pair selection affects performance and whether \mymethodFON{} preserves the neuron-specific input differentiation that motivates our approach.

\begin{figure}[h]
    \centering 
    \includegraphics[width=1.0\linewidth]{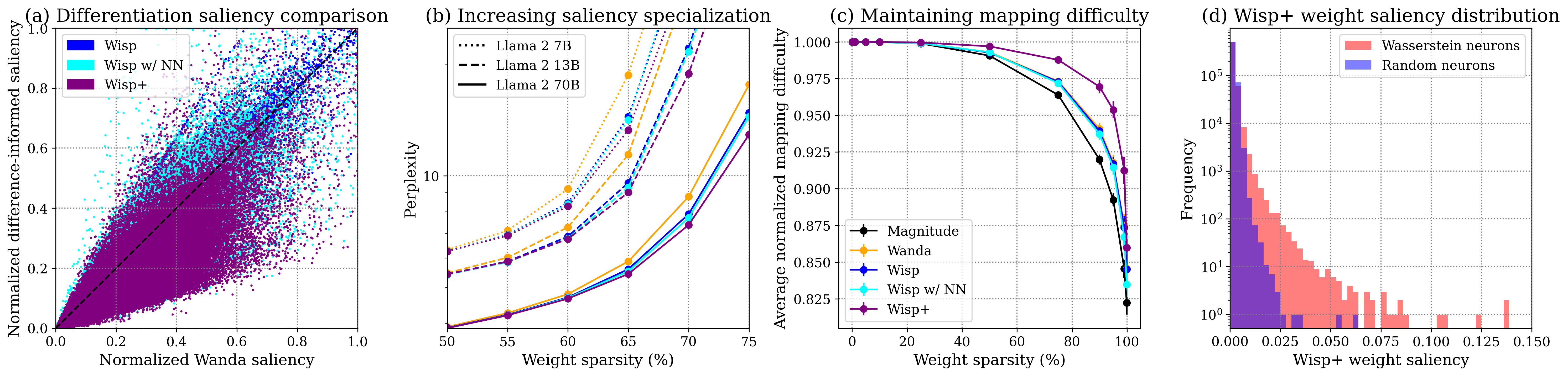}
    \caption{Analysis of difference-informed pruning and input differentiation. Saliency comparison of \mymethodFOL{}, nearest-neighbor \mymethodFOL{}, and \mymethodFON{} against Wanda (a). Perplexity in the Llama 2 family as pair selection becomes more targeted (b). Preservation of mapping difficulty (MD) in Wasserstein neurons under increasing sparsity (c). \mymethodFON{} weight saliency distributions for Wasserstein versus randomly chosen neurons (d). (a,c,d) Analysis performed in second gate projection of Llama 3.1 8B. MD is used only for diagnostic analysis, not pruning. Error bars indicate one standard error of the mean.}
    \label{fig:wisp_analysis}
\end{figure}

First, the difference-informed saliencies remain correlated with Wanda due to their shared dependence on weight magnitude, but exhibit substantial off-diagonal variation (\Cref{fig:wisp_analysis}a). We next examine whether input similarity itself contributes to the pruning signal by comparing \mymethodFOL{} with $\Delta\mathbf{X}$ formed from random and from nearest-neighbor pairing. The latter consistently improves perplexity in more constrained settings, with \mymethodFON{} generally improving further through neuron-specific, output-aware selection (\Cref{fig:wisp_analysis}b). These results reinforce the motivating role of input differentiation, although more targeted pair selection increases pruning time. We therefore retain random-pair \mymethodFOL{} as the primary layerwise first-order method, as it captures much of the benefit at lower cost (\Cref{sec:sampling}).

We next investigate whether these criteria preserve the motivating neuron-specific behavior. Following \cite{kong2025negative}, we define Wasserstein neurons as those with the top 1\% mapping difficulty (MD), which measures their ability to separate similar inputs. We use the formulation established in prior work, with implementation details in \Cref{sec:mapping_diff}. Under pruning, \mymethodFON{} preserves MD most effectively, particularly at high sparsity (\Cref{fig:wisp_analysis}c). In contrast, \mymethodFOL{} and its nearest-neighbor variant track Wanda more closely, since their shared layerwise statistics do not explicitly target the separations performed by individual neurons. Thus, improvements in overall perplexity need not translate directly into preservation of this neuron-specific diagnostic.

Finally, we examine whether \mymethodFON{} localizes saliency to weights in Wasserstein neurons. Compared to an equal number of randomly selected neurons, Wasserstein neurons exhibit a substantially heavier tail of high-saliency weights (\Cref{fig:wisp_analysis}d). While prior work used Wasserstein distance and MD to identify these neurons \citep{sawmya2025wasserstein}, \mymethodFON{} provides weight-level saliency associated with their behavior. Together, these results show that \mymethodFON{} not only improves pruning performance but also better preserves the motivating phenomenon, without using MD directly in its pruning criterion.

\subsection{Component-wise comparisons}

We validate where difference-informed pruning is most useful. For each component type, we prune only that component in Llama 2 7B at 2:4 sparsity and compare the corresponding standard and difference-informed criteria in terms of final perplexity. The gains are concentrated in the MLP gate and up projections, especially when both are pruned together. This matches the motivation from Wasserstein neurons, which are most prominently observed there \citep{sawmya2025wasserstein}. Other components are less consistent: the MLP down projection worsens under difference-informed saliency, and attention shows mixed behavior, with a small gain for \mymethodFON{} but not for \mymethodFOL{} or \mymethodSO{}. We therefore use the difference-informed criteria only for the MLP gate and up projections throughout the paper, and retain the corresponding baseline method for attention and down projections.

\begin{table}[h]
\centering
\begin{tabular}{lcccccc}
\toprule
Component & \% of all parameters & Wanda & \mymethodFOL{} & \mymethodFON{} & SparseGPT & \mymethodSO{}\\
\midrule
MLP gate & 22.3\% & 5.8410 & 5.8162 & 5.7224 & 5.4557 & 5.4203\\
MLP up & 22.3\% & 5.5759 & 5.5704 & 5.5724 & 5.4686 & 5.4696\\
MLP gate + up & 44.6\% & 6.7854 & 6.6412 & 6.5035 & 6.0176 & 5.9018\\
MLP down & 22.3\% & 5.6289 & 5.6762 & 5.7318 & 5.5273 & 5.5414\\
Attention & 33.1\% & 6.7038 & 6.7617 & 6.6270 & 5.7504 & 5.7743\\
\midrule
\end{tabular}
\caption{Component-wise comparison of standard and difference-informed pruning criteria. Values report WikiText-2 test perplexity. Results are averaged over three trials.}
\label{tab:component_analysis}
\end{table}

\subsection{Generalization to Mistral and Qwen}

We next evaluate whether the gains generalize beyond the Llama families. \Cref{tab:mistral_qwen} reports 2:4 sparsity results for Mistral 7B v0.3, Qwen3 8B Base, and Qwen3 14B Base.

\begin{table}[h]
\centering
\begin{tabular}{lcccccccc}
\toprule
& \multicolumn{2}{c}{\underline{Mistral 7B v0.3}} & \multicolumn{2}{c}{\underline{Qwen3 8B Base}} & \multicolumn{2}{c}{\underline{Qwen3 14B Base}}\\

Method & PPL & Mean & PPL & Mean & PPL & Mean \\
\midrule
Dense & $4.95$ & $60.56$ & $6.51$ & $68.25$ & $5.95$ & $71.70$ \\
\midrule
Magnitude & $13.49$ & $42.12$ & $754.4$ & $31.92$ & $121.3$ & $41.57$ \\
Wanda & $9.64$ & $41.49$ & $11.33$ & $50.31$ & $8.81$ & $56.66$ \\
\mymethodFOL{} & $8.97$ & $42.74$ & $11.18$ & $\underline{50.49}$ & $8.68$ & $56.65$ \\
\mymethodFON{} & $\underline{8.64}$ & $\underline{43.05}$ & $\underline{10.86}$ & $50.46$ & $\underline{8.55}$ & $\underline{57.03}$ \\
SparseGPT & $7.47$ & $44.79$ & $9.04$ & $51.12$ & $7.80$ & $57.82$ \\
\mymethodSO{} & $\mathbf{6.91}$ & $\mathbf{45.97}$ & $\mathbf{8.82}$ & $\mathbf{51.45}$ & $\mathbf{7.65}$ & $\mathbf{58.11}$ \\
\midrule
\end{tabular}
\caption{Performance on Mistral 7B v0.3, Qwen3 8B Base, and Qwen3 14B Base at 2:4 sparsity. Bold indicates best performance and underline indicates best performance without weight updates. Results collected over three trials.}
\label{tab:mistral_qwen}
\end{table}

These results show that the benefit of preserving input differentiation is not specific to Llama 2 or Llama 3.1. The improvements transfer across architectures and model families, including both Mistral and Qwen models. This supports the view that pairwise input separation provides a general calibration signal for post-training sparsification. Additional results on Granite 4.1 in \Cref{sec:granite_ppl}.

\subsection{Composability with stronger pruning methods}
\label{sec:composability}

We next test whether difference-informed pruning composes with stronger methods. RIA augments Wanda-style pruning with row and column normalized weight importance and modified activation scaling. We form RIA-\mymethodFOL{} and RIA-\mymethodFON{} by retaining RIA’s weight normalization while replacing its activation statistic in the MLP gate and up projections with the corresponding \mymethodFOL{} or \mymethodFON{} difference statistic. Similarly, ALPS improves second-order pruning through a stronger sparse-reconstruction optimizer. We form ALPS-\mymethodSO{} by retaining the complete ALPS procedure and hyperparameters while replacing its standard Hessian in the gate and up projections with the \mymethodSO{} regularized difference-informed Hessian, using the same $\gamma=0.01$. More details in \Cref{sec:composability_full}.

\begin{table}[h]
\centering
\begin{tabular}{lccccccccc}
\toprule
& \multicolumn{3}{c}{\underline{Llama 2 7B}} & \multicolumn{3}{c}{\underline{Llama 2 13B}} & \multicolumn{3}{c}{\underline{Llama 2 70B}} \\
Method & 2:4 & 50\% & 65\% & 2:4 & 50\% & 65\% & 2:4 & 50\% & 65\% \\
\midrule
RIA & 10.24 & 6.25 & 18.23 & 7.48 & 5.39 & 10.22 & 4.91 & 3.84 & 5.79\\
RIA-\mymethodFOL{} & 9.57 & 6.22 & 15.83 & 7.19 & 5.38 & 9.30 & 4.85 & \underline{3.83} & 5.59\\
RIA-\mymethodFON{} & \underline{9.34} & \underline{6.21} & \underline{14.52} & \underline{7.13} & \underline{5.37} & \underline{8.94} & \underline{4.81} & \underline{3.83} & \underline{5.46}\\
ALPS & 7.22 & \textbf{5.88} & 8.24 & 6.11 & 5.17 & 6.80 & 4.50 & 3.74 & 4.80\\
ALPS-\mymethodSO{} & \textbf{7.04} & 5.89 & \textbf{7.94} & \textbf{5.97} & \textbf{5.15} & \textbf{6.54} & \textbf{4.41} & \textbf{3.72} & \textbf{4.66}\\

\midrule
\end{tabular}
\caption{Composability with stronger pruning methods on WikiText-2. Difference-informed variants improve RIA in all settings and ALPS in all but one. Bold indicates best performance and underline indicates best performance without weight updates. Results are averaged over three trials.}
\label{tab:composability}
\end{table}

\begin{figure}[h]
    \centering 
    \includegraphics[width=1\linewidth]{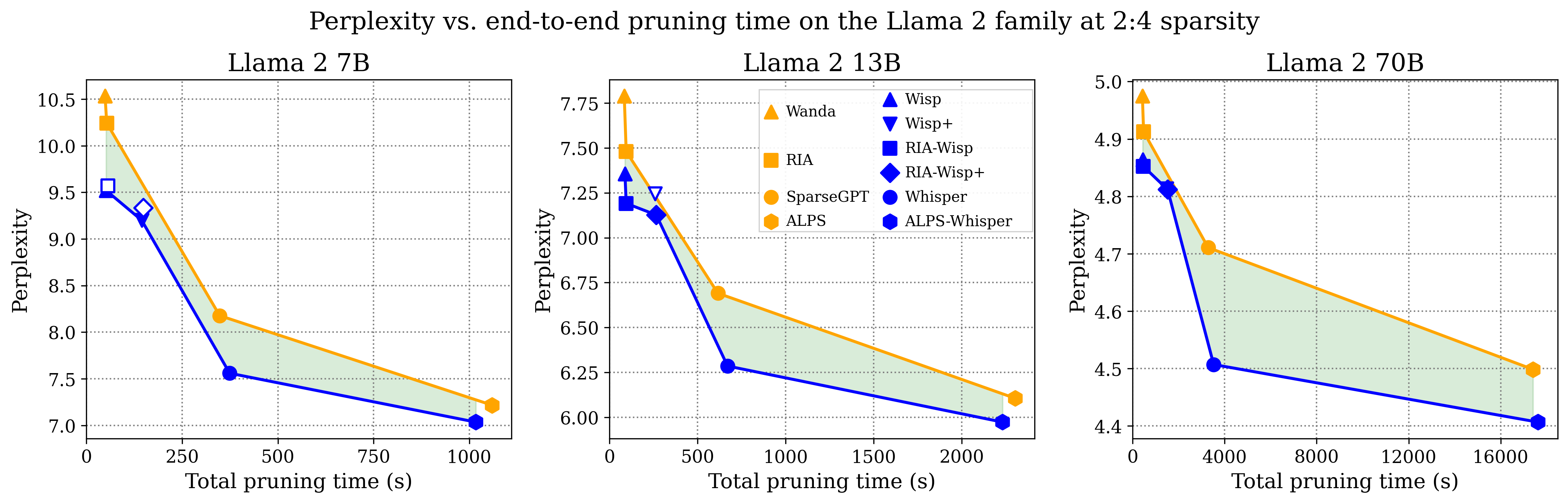}
    \caption{Difference-informed pruning shifts the accuracy-runtime frontier outward. Perplexity vs. end-to-end pruning time is shown for the Llama 2 family at 2:4 sparsity (lower-left is better). Orange markers denote baseline pruning methods, while blue markers denote their difference-informed counterparts. Filled markers denote Pareto-efficient methods and hollow markers denote dominated methods. Lines connect Pareto-efficient points within each group, and the shaded region highlights the improvement in attainable accuracy-runtime tradeoffs. Results collected over three trials.}
    \label{fig:pareto}
\end{figure}

Across Llama 2 7B, 13B, and 70B at 2:4, 50\%, and 65\% sparsity, RIA-\mymethodFOL{} improves upon RIA in all nine settings with negligible pruning time overhead. RIA-\mymethodFON{} improves further, showing that difference-informed saliency is complementary to RIA’s weight normalization. ALPS-\mymethodSO{} similarly improves ALPS in eight of nine settings with at most 1.5\% additional pruning time; the sole exception differs by only 0.01 perplexity for Llama 2 7B at 50\% sparsity. \Cref{fig:pareto} illustrates the resulting accuracy-runtime tradeoff at 2:4 sparsity. In particular, on Llama 2 70B, \mymethodSO{} nearly matches ALPS (4.51 compared to 4.50 perplexity) while requiring only 20\% total pruning time, and ALPS-\mymethodSO{} further reduces perplexity to 4.41 at essentially the same cost as ALPS. Thus, difference-informed criteria are not specific to Wanda or SparseGPT, but improve stronger pruning formulations and shift the overall accuracy-runtime frontier. Full results provided in \Cref{sec:composability_tables}.

\section{Limitations}
\label{sec:limitations}

Our experiments focus on dense pretrained language models and one-shot pruning without subsequent finetuning. This is the standard setting for many LLM sparsity studies, since dense models provide a controlled benchmark for comparing pruning criteria without additional architectural or systems complications. Extending the method to mixture of experts (MoE) models is an interesting direction, but MoE sparsification generally requires more engineering around routing and load balancing. Because MoE architectures allocate a larger fraction of parameters to expert MLP up and gate projections, difference-informed pruning may be especially useful there. Similarly, our techniques only modify the saliencies of the up and gate projections, and we do not propose changes to the down or attention projections. Despite this restriction, we observe consistent performance. We also do not study post-pruning finetuning or recovery training. Finetuning could improve all pruning methods, but it would also act as a confound. We therefore isolate the pure one-shot setting, where differences between methods can be attributed directly to the pruning criterion or reconstruction geometry.

\section{Conclusion and Discussion}

Across sparsity patterns, model scales, model families, and pruning formulations, our results show that one-shot pruning benefits from preserving pairwise output differences. \mymethodFON{} demonstrates that accounting for how individual neurons differentiate similar inputs improves pruning, while also better preserving this motivating behavior. \mymethodFOL{} shows that a simpler layerwise difference statistic also improves Wanda at essentially the same cost. \mymethodSO{} extends global difference preservation to second-order pruning by replacing SparseGPT's activation Hessian with a lightly regularized difference-informed Hessian. Across our primary Wanda and SparseGPT comparisons, \mymethodSO{} obtains the lowest perplexity in every pruned setting and improves downstream averages across Llama, Mistral, and Qwen models. Composing these criteria with RIA and ALPS shifts the accuracy-runtime frontier outward: RIA-\mymethodFOL{} and RIA-\mymethodFON{} improve RIA and ALPS-\mymethodSO{} improves ALPS in nearly every setting with negligible additional pruning time. Together, these results establish difference-informed pruning as a broadly effective principle for one-shot LLM sparsification, improving performance across backbones, saliency formulations, and reconstruction solvers.

More broadly, our findings suggest that pruning criteria may benefit by being tailored to the computation performed by each model component. The MLP up and gate projections appear to rely on maintaining differences between outputs, and preserving this behavior improves sparsification. Other components may require different preservation principles. We hope this work motivates the development of component-specific pruning criteria that preserve not only large activations or reconstruction fidelity, but also the functional role each component plays in the model.

\clearpage

\section*{Acknowledgments}

The authors would like to thank Nima Dehmamy, Daniel Weidele, Eldar Kurtić, Alexandre Marques, Rebecca Lin, Timothy Gomez, and Shashata Sawmya for providing insightful feedback and constructive discussions during the development of this work.

This project was supported by an MIT-IBM Watson AI Lab grant, by Jane Street, by Lightspeed, and by the MIT UROP Office. Computational resources were provided by the MIT Office of Research Computing and Data.

\clearpage
\bibliographystyle{plainnat}
\bibliography{neurips_2026}
\clearpage


\appendix

\section{Appendix}

\subsection{Difference preservation objective}
\label{sec:difference_derivation_whisper}

Consider a linear layer with dense weights $\mathbf{W}$ and sparse weights
$\widehat{\mathbf{W}}$. Standard layerwise reconstruction methods such as
SparseGPT aim to preserve the layer output on calibration inputs:
\[
\min_{\widehat{\mathbf{W}}}
\left\|
\mathbf{W}\mathbf{X}
-
\widehat{\mathbf{W}}\mathbf{X}
\right\|_F^2
\]
Writing $\mathbf{E}=\mathbf{W}-\widehat{\mathbf{W}}$, this becomes
\[
\|\mathbf{E}\mathbf{X}\|_F^2
=
\mathrm{tr}(\mathbf{E}\mathbf{X}\mathbf{X}^{T}\mathbf{E}^{T})
\]
Letting $\mathbf{e}_i$ denote row $i$ of $\mathbf{E}$, we can equivalently
write this objective as
\[
\|\mathbf{E}\mathbf{X}\|_F^2
=
\sum_{i=1}^{n}
\mathbf{e}_i
\mathbf{X}\mathbf{X}^{T}
\mathbf{e}_i^{T}
\]
Thus, for each output row, the reconstruction objective is a quadratic form
whose curvature matrix is:
\[
\mathbf{H}=\mathbf{X}\mathbf{X}^{T}
\]

Our goal is instead to preserve the layer's ability to separate paired
calibration inputs. Let
\[
\Delta\mathbf{X}=\mathbf{X}-\mathbf{X}_{\pi}
\]
be the matrix of sampled pairwise differences. The corresponding objective is
\[
\min_{\widehat{\mathbf{W}}}
\left\|
\mathbf{W}\Delta\mathbf{X}
-
\widehat{\mathbf{W}}\Delta\mathbf{X}
\right\|_F^2
=
\min_{\widehat{\mathbf{W}}}
\|\mathbf{E}\Delta\mathbf{X}\|_F^2
\]
Expanding gives
\[
\|\mathbf{E}\Delta\mathbf{X}\|_F^2
=
\mathrm{tr}
\left(
\mathbf{E}
\Delta\mathbf{X}\Delta\mathbf{X}^{T}
\mathbf{E}^{T}
\right)
=
\sum_{i=1}^{n}
\mathbf{e}_i
\Delta\mathbf{X}\Delta\mathbf{X}^{T}
\mathbf{e}_i^{T}
\]
Thus, preserving pairwise output differences yields the difference curvature
matrix
\[
\mathbf{H}_{\Delta}
=
\Delta\mathbf{X}\Delta\mathbf{X}^{T}
\]
This, with the addition of light regularization, is the Hessian used by
\mymethodSO{}.

Since \mymethodSO{} otherwise follows the SparseGPT update rule, replacing
$\mathbf{H}$ by the regularized difference Hessian $
\widetilde{\mathbf{H}}
=
\gamma \mathbf{H}
+
(1-\gamma)
\frac{\mu}{\mu_{\Delta}}
\mathbf{H}_{\Delta}
$
directly yields the score
\[
\mathbf{S}_{ij}^{\text{\mymethodSO}}
=
\frac{|\mathbf{W}_{ij}|^2}
{
\left[
(\widetilde{\mathbf{H}}+\lambda\mathbf{I})^{-1}
\right]_{jj}
}
\]

\clearpage

\subsection{Diagonal approximation and \mymethodFOL{}}
\label{sec:difference_derivation_wisp}

The same difference-preservation objective also yields the first-order
\mymethodFOL{} saliency under a diagonal approximation to the curvature. For
one output row $i$, the difference-preserving reconstruction error is
\[
\mathbf{e}_i
\mathbf{H}_{\Delta}
\mathbf{e}_i^T
\]
with the Hessian of differences
\[
\mathbf{H}_{\Delta}
=
\Delta\mathbf{X}\Delta\mathbf{X}^T
\]
where $\mathbf{e}_i=\mathbf{w}_i-\widehat{\mathbf{w}}_i$. If we ignore
cross-channel interactions by approximating
\[
\mathbf{H}_{\Delta}
\approx
\operatorname{diag}(\mathbf{H}_{\Delta})
\]
then the row-wise objective becomes
\[
\sum_{j=1}^{m}
e_{ij}^{2}
(\mathbf{H}_{\Delta})_{jj}
\]
Since
\[
(\mathbf{H}_{\Delta})_{jj}
=
\sum_{\ell=1}^{s}
(\Delta\mathbf{X}_{j,\ell})^2
=
\|\Delta\mathbf{X}_{j}\|_2^2
\]
the diagonal reconstruction error is
\[
\sum_{j=1}^{m}
(\mathbf{W}_{ij}-\widehat{\mathbf{W}}_{ij})^2
\|\Delta\mathbf{X}_{j}\|_2^2
\]
In a no-update pruning rule, removing weight $\mathbf{W}_{ij}$ sets
$\widehat{\mathbf{W}}_{ij}=0$, and therefore incurs diagonal cost
\[
\mathbf{W}_{ij}^{2}
\|\Delta\mathbf{X}_{j}\|_2^2
\]
Ranking weights by this cost is equivalent to ranking by its square root,
\[
|\mathbf{W}_{ij}|
\|\Delta\mathbf{X}_{j}\|_2
\]
which is precisely the \mymethodFOL{} saliency. This is directly analogous to
the diagonal reconstruction view of Wanda, with the activation norm
$\|\mathbf{X}_{j}\|_2$ replaced by the difference norm
$\|\Delta\mathbf{X}_{j}\|_2$.

\clearpage 

\subsection{Sampling approach details}
\label{sec:sampling}

To test whether additional samples of pairwise differences improves performance, we increase the number of independent random permutations to obtain $N \in \{s, 2s, 4s, 8s, 16s\}$ pairs per calibration sequence of length $s$, keeping $K=0.005N$ for \mymethodFON{}. 

\begin{table}[h]
\centering
\begin{tabular}{lccccc}
\toprule
& \multicolumn{5}{c}{\underline{Sampled pairs per sequence}}\\
Method & $s$ & $2s$& $4s$& $8s$& $16s$\\
\midrule
\mymethodFOL{} & 9.52 & 9.54 & 9.52 & 9.54 & 9.52\\
\mymethodFON{} & 9.20 & 9.21 & 9.21 & 9.20 & 9.23\\
\mymethodSO{} & 7.56 &7.57 & 7.55&7.56&7.54\\ 
\midrule
\end{tabular}
\caption{One permutation per calibration sequence is sufficient in practice. We report WikiText-2 perplexity for Llama 2 7B at 2:4 sparsity. Results are averaged over three trials.}
\label{tab:pair_number}
\end{table}

Performance remains essentially unchanged across all pair counts for \mymethodFOL{}, \mymethodFON{}, and \mymethodSO{}, indicating that one permutation ($N=s$) is sufficient in practice and more computationally efficient.

Furthermore, to directly isolate whether input similarity itself contributes to the pruning signal, even without Wisp+'s neuron-specific saliency, we run the following nearest neighbor pairing ablation for \mymethodFOL{}. For a calibration sequence containing activation vectors $\mathbf{x}_1, \mathbf{x}_2,\ldots \mathbf{x}_s$, rather than forming $\Delta \mathbf{X}$ using a random permutation to create pairs as in \Cref{sec:wisp}, we now directly find the nearest neighbor of each $\mathbf{x}_i$ via the $L_2$ norm: $n(i)=\text{argmin}_{j\in\{1,2,\ldots s\}, j\neq i} ||\mathbf{x}_i - \mathbf{x}_j||_2$.

We then form $\Delta \mathbf{x}_i^{NN} = \mathbf{x}_i - \mathbf{x}_{n(i)}$ and thus construct $\Delta \mathbf{X}_{NN} = [\mathbf{x}_1 - \mathbf{x}_{n(1)}, \mathbf{x}_2 - \mathbf{x}_{n(2)},\ldots \mathbf{x}_s - \mathbf{x}_{n(s)}]$. This preserves the same number of pairwise differences as the random-permutation implementation. All other aspects of \mymethodFOL{} and the evaluation protocol remain unchanged for this diagnostic ablation. We report WikiText-2 perplexity averaged over three trials.

\begin{table}[h]
\centering
\resizebox{\linewidth}{!}{
\begin{tabular}{lccccccccc}
\toprule
& \multicolumn{3}{c}{Llama 2 7B} & \multicolumn{3}{c}{Llama 2 13B} & \multicolumn{3}{c}{Llama 2 70B} \\
\cmidrule(lr){2-4} \cmidrule(lr){5-7} \cmidrule(lr){8-10}
Method & 2:4 & 50\% & 65\% & 2:4 & 50\% & 65\% & 2:4 & 50\% & 65\% \\
\midrule
\mymethodFOL{} with random pairs&9.52&6.25&14.61&7.36&5.42&9.59&4.86&3.89&5.60 \\
\mymethodFOL{} with nearest neighbors&9.42&6.26&14.14&7.27&5.41&9.33&4.84&3.88&5.53 \\ 
\mymethodFON{} &9.20&6.25&13.17&7.24&5.42&9.04&4.81&3.88&5.44 \\
\midrule
\end{tabular}
}
\caption{Nearest neighbor pairing ablation on the Llama 2 family. Replacing random pairs in \mymethodFOL{} with nearest neighbor input activation pairs consistently improves perplexity in the more constrained 2:4 and 65\% sparsity settings, while performance remains similar at 50\% sparsity. \mymethodFON{} generally improves further through neuron-specific, output-aware pair selection. Results are averaged over three trials.}
\label{tab:nearest_neighbors}
\end{table}

\begin{table}[h]
\centering
\begin{tabular}{lccc}
\toprule
Method & Llama 2 7B & Llama 2 13B & Llama 2 70B \\
\midrule
\mymethodFOL{} with random pairs&51.3 s&87.3 s&434 s\\
\mymethodFOL{} with nearest neighbors&84.0 s&139 s&573 s\\ 
\mymethodFON{} &144 s&258 s&1476 s\\
\midrule
\end{tabular}
\caption{More targeted pair selection improves performance at increased pruning cost. Results are averaged over three trials.}
\label{tab:nearest_neighbors_time}
\end{table}

Nearest neighbor pairing improves \mymethodFOL{} at 2:4 and 65\% sparsity while remaining comparable at 50\%, showing that emphasizing differences between similar inputs strengthens the layerwise statistic in more constrained settings. However, this incurs a 30\% to 65\% increase in pruning time; we therefore do not use nearest neighbor pairing in our primary experiments. \mymethodFON{} generally improves further through neuron-specific, output-aware pair selection. Overall, these results support the difference-informed framework: the progression from global difference preservation to increasingly targeted preservation of input separation improves performance, with a corresponding tradeoff in computational cost.

\clearpage

\subsection{Centering interpretation of random pairing}
\label{sec:centering}

The random-pair statistics used by \mymethodFOL{} and \mymethodSO{} admit an
equivalent centered interpretation at the second-moment level. Let
\[
\bar{\mathbf{x}}
=
\frac{1}{s}\mathbf{X}\mathbf{1}
\]
denote the mean calibration input and define the centered calibration matrix
\[
\mathbf{X}_c
=
\mathbf{X}
-
\bar{\mathbf{x}}\mathbf{1}^{T}
\]
Recall that $\Delta\mathbf{X}
=
\mathbf{X}-\mathbf{X}_{\pi}
$,
where $\pi$ is a uniformly sampled random permutation of the calibration
inputs. Taking the expectation over $\pi$ gives
\begin{align*}
\mathbb{E}_{\pi}
\left[
\Delta\mathbf{X}\Delta\mathbf{X}^{T}
\right]
&=
\mathbb{E}_{\pi}
\left[
(\mathbf{X}-\mathbf{X}_{\pi})
(\mathbf{X}-\mathbf{X}_{\pi})^{T}
\right] \\
&=
\mathbf{X}\mathbf{X}^{T}
+
\mathbb{E}_{\pi}
\left[
\mathbf{X}_{\pi}\mathbf{X}_{\pi}^{T}
\right]
-
\mathbb{E}_{\pi}
\left[
\mathbf{X}\mathbf{X}_{\pi}^{T}
\right]
-
\mathbb{E}_{\pi}
\left[
\mathbf{X}_{\pi}\mathbf{X}^{T}
\right]
\end{align*}
Permuting the columns does not change their second moment, so
\[
\mathbf{X}_{\pi}\mathbf{X}_{\pi}^{T}
=
\mathbf{X}\mathbf{X}^{T}
\]
Moreover, each permuted column is uniformly distributed over the columns of
$\mathbf{X}$, and therefore
\[
\mathbb{E}_{\pi}
\left[
\mathbf{X}\mathbf{X}_{\pi}^{T}
\right]
=
\mathbb{E}_{\pi}
\left[
\mathbf{X}_{\pi}\mathbf{X}^{T}
\right]
=
s\bar{\mathbf{x}}\bar{\mathbf{x}}^{T}
\]
It follows that
\[
\mathbb{E}_{\pi}
\left[
\Delta\mathbf{X}\Delta\mathbf{X}^{T}
\right]
=
2\left(
\mathbf{X}\mathbf{X}^{T}
-
s\bar{\mathbf{x}}\bar{\mathbf{x}}^{T}
\right) \\
=
2\mathbf{X}_c\mathbf{X}_c^{T}
\]
Thus, the random-pair difference Hessian is a stochastic estimator of twice
the centered second-moment matrix.

For \mymethodFOL{}, taking the diagonal gives
\[
\mathbb{E}_{\pi}
\left[
\|\Delta\mathbf{X}_{j}\|_2^2
\right]
=
2\|\mathbf{X}_{c,j}\|_2^2
\]
The corresponding expected diagonal pruning cost is therefore
\[
2\mathbf{W}_{ij}^{2}
\|\mathbf{X}_{c,j}\|_2^2
\]
Ranking weights by the square root of this cost yields
\[
\sqrt{2}
|\mathbf{W}_{ij}|
\|\mathbf{X}_{c,j}\|_2
\]
which has the same ordering as the deterministic centered saliency
\[
\mathbf{S}_{ij}^{\text{Centered \mymethodFOL{}}}
=
|\mathbf{W}_{ij}|
\|\mathbf{X}_{c,j}\|_2
\]
The common factor $\sqrt{2}$ does not affect pruning decisions.

The same equivalence holds for \mymethodSO{}. Let
\[
\mathbf{H}_c
=
\mathbf{X}_c\mathbf{X}_c^{T},
\qquad
\mu_c
=
\frac{\operatorname{tr}(\mathbf{H}_c)}{m}
\]
Since the expected random-pair Hessian is $2\mathbf{H}_c$, its trace scale is
also multiplied by two. Consequently, the factor cancels under the
normalization used by \mymethodSO{}:
\[
\frac{\mu}{2\mu_c}
\left(2\mathbf{H}_c\right)
=
\frac{\mu}{\mu_c}\mathbf{H}_c
\]
The deterministic centered form of the regularized Hessian is therefore
\[
\widetilde{\mathbf{H}}_c
=
\gamma\mathbf{H}
+
(1-\gamma)
\frac{\mu}{\mu_c}
\mathbf{H}_c
\]
which has the same normalized geometry as the expected random-pair
construction.

Hence, random pairing and direct centering represent the same global
second-moment statistic up to a constant factor: random pairing estimates it
stochastically, whereas centering computes it deterministically. This identity
also holds when the statistics are constructed separately within each
calibration sequence and subsequently aggregated.

To assess whether the two formulations also match in practice, we compare their pruning performance across Llama 2 scales and sparsity settings.

\begin{table}[H]
\centering
\begin{tabular}{lccccccccc}
\toprule
& \multicolumn{3}{c}{Llama 2 7B} & \multicolumn{3}{c}{Llama 2 13B} & \multicolumn{3}{c}{Llama 2 70B} \\
\cmidrule(lr){2-4} \cmidrule(lr){5-7} \cmidrule(lr){8-10}
Method & 2:4 & 50\% & 65\% & 2:4 & 50\% & 65\% & 2:4 & 50\% & 65\% \\
\midrule
\mymethodFOL{}&9.52&6.25&14.61&7.36&5.42&9.59&4.86&3.89&5.60 \\
Centered \mymethodFOL{} & 9.53&6.25&14.56 & 7.36&5.42&9.58 & 4.86&3.89&5.60 \\
\mymethodSO{} &7.56 & 6.04&9.46 & 6.29&5.28&7.54 & 4.51&3.77&4.83\\
Centered \mymethodSO{} & 7.56&6.04&9.45 & 6.28&5.28&7.51 & 4.50&3.77&4.82 \\
\midrule
\end{tabular}
\caption{Centered implementations closely match the random-pair formulations of \mymethodFOL{} and \mymethodSO{} across Llama 2 scales and sparsities. Results are averaged over three trials.}
\label{tab:centering}
\end{table}

The centered variants closely match the random-pair formulations because they deterministically compute the same second-moment geometry, up to a common factor that does not affect pruning. We retain pairwise output-difference preservation as the conceptual framework because explicit pairs naturally support both \mymethodFON{}'s neuron-specific selection and similarity-focused constructions such as the nearest neighbor pairing ablation in \Cref{tab:nearest_neighbors}. Accordingly, the primary experiments use the random-pair formulation, while the centered variants provide deterministic implementations of the global \mymethodFOL{} and \mymethodSO{} statistics. We provide both implementations in the code.

\clearpage

\subsection{Regularizing $\mathbf{H}_\Delta$ with $\gamma$}
\label{sec:gamma_justification}
In practice, using only $\mathbf{H}_{\Delta}$ may be ill-conditioned and may
overemphasize preserving differences at the expense of preserving absolute
outputs. We therefore combine the difference Hessian with the standard
activation Hessian:
\[
\widetilde{\mathbf{H}}
=
\gamma \mathbf{H}
+
(1-\gamma)
\frac{\mu}{\mu_{\Delta}}
\mathbf{H}_{\Delta}
\]
where 
$
\mu=\frac{\operatorname{tr}(\mathbf{H})}{m}$ and $
\mu_{\Delta}=\frac{\operatorname{tr}(\mathbf{H}_{\Delta})}{m}$. This corresponds to the
regularized objective
\[
\gamma
\|\mathbf{E}{\mathbf{X}}\|_F^2
+
(1-\gamma)\frac{\mu}{\mu_{\Delta}}
\|\mathbf{E}\Delta{\mathbf{X}}\|_F^2
\]
which preserves pairwise separations while retaining a small amount of the
standard reconstruction objective. We conduct experiments for the choice of $\gamma$ below, both on the level of an individual layer and end to end. Note that $\gamma = 1$ is equivalent to using the baseline SparseGPT algorithm.

\begin{figure}[h]
    \centering 
    \includegraphics[width=0.9\linewidth]{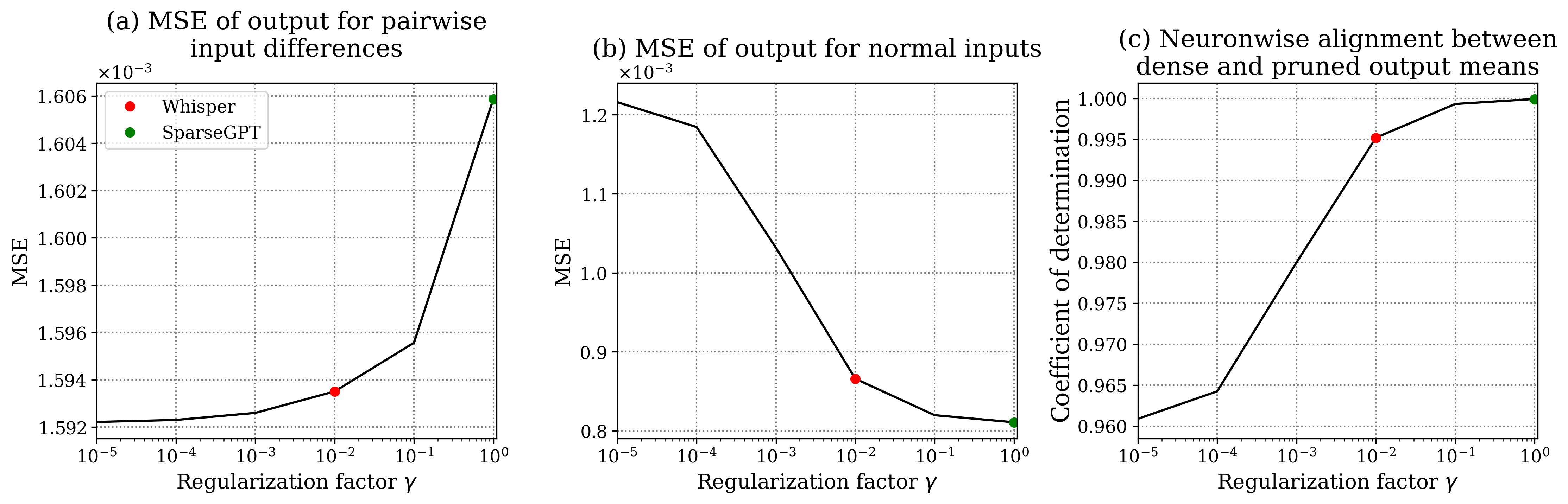}
    \caption{Effect of $\gamma$ for an individual layer. Reconstruction error for the outputs of the pairwise input differences (a). Reconstruction error for the regular outputs (b). Coefficient of determination of the neuronwise sparse output means compared to the neuronwise dense output means (c). The $\gamma$ used for all experiments in the main paper, 0.01, is marked, as well as the effective $\gamma$ for SparseGPT of 1. Analysis performed in the second gate projection of Llama 3.1 8B. The layer is pruned to 2:4 sparsity.}
    \label{fig:gamma_explanation}
\end{figure}

First, we evaluate the reconstruction error for both regular outputs $\mathbf{W}\mathbf{X}$ and the outputs of pairwise differences of inputs $\mathbf{W}\Delta\mathbf{X}$ for different values of $\gamma$ in \mymethodSO{} at 2:4 sparsity. We find that using smaller $\gamma$ minimizes the mean square error (MSE) of reconstructing $\mathbf{W}\Delta\mathbf{X}$ with the sparse weights $\widehat{\mathbf{W}}$, with $\gamma = 1$ having the highest MSE (\Cref{fig:gamma_explanation}a). However, this comes at the cost of worse reconstruction of $\mathbf{W}\mathbf{X}$ itself, with decreasing MSE for $\mathbf{W}\mathbf{X}$ as $\gamma$ increases (\Cref{fig:gamma_explanation}b).

When we compute the mean of the outputs for each neuron $i$, $\operatornamewithlimits{mean}\mathbf{w}_i\mathbf{X}$, and compare them to the mean of the outputs under sparsity, $\operatornamewithlimits{mean}\widehat{\mathbf{w}}_i\mathbf{X}$, we observe that lower $\gamma$ causes a greater shift in the means in favor of preserving the differences (\Cref{fig:gamma_explanation}c).

Overall, this suggests that, while using a smaller $\gamma$ will better preserve outputs of pairwise input differences, a $\gamma$ that is too small sacrifices reconstruction of the regular outputs and systematically misaligns the neuronwise means, ultimately causing worse end to end recovery. We thus choose $\gamma = 0.01$ for all experiments in the main paper, which has a balance of low MSE for the outputs of pairwise differences of inputs, low MSE for the outputs of regular inputs, and high alignment between neuronwise dense output means and neuronwise pruned output means.

\clearpage

We also measure end to end perplexity when sparsifying individual components in each transformer block, with different $\gamma$ values. We do so in Llama 2 7B at 2:4 sparsity.

\begin{figure}[h]
    \centering 
    \includegraphics[width=0.9\linewidth]{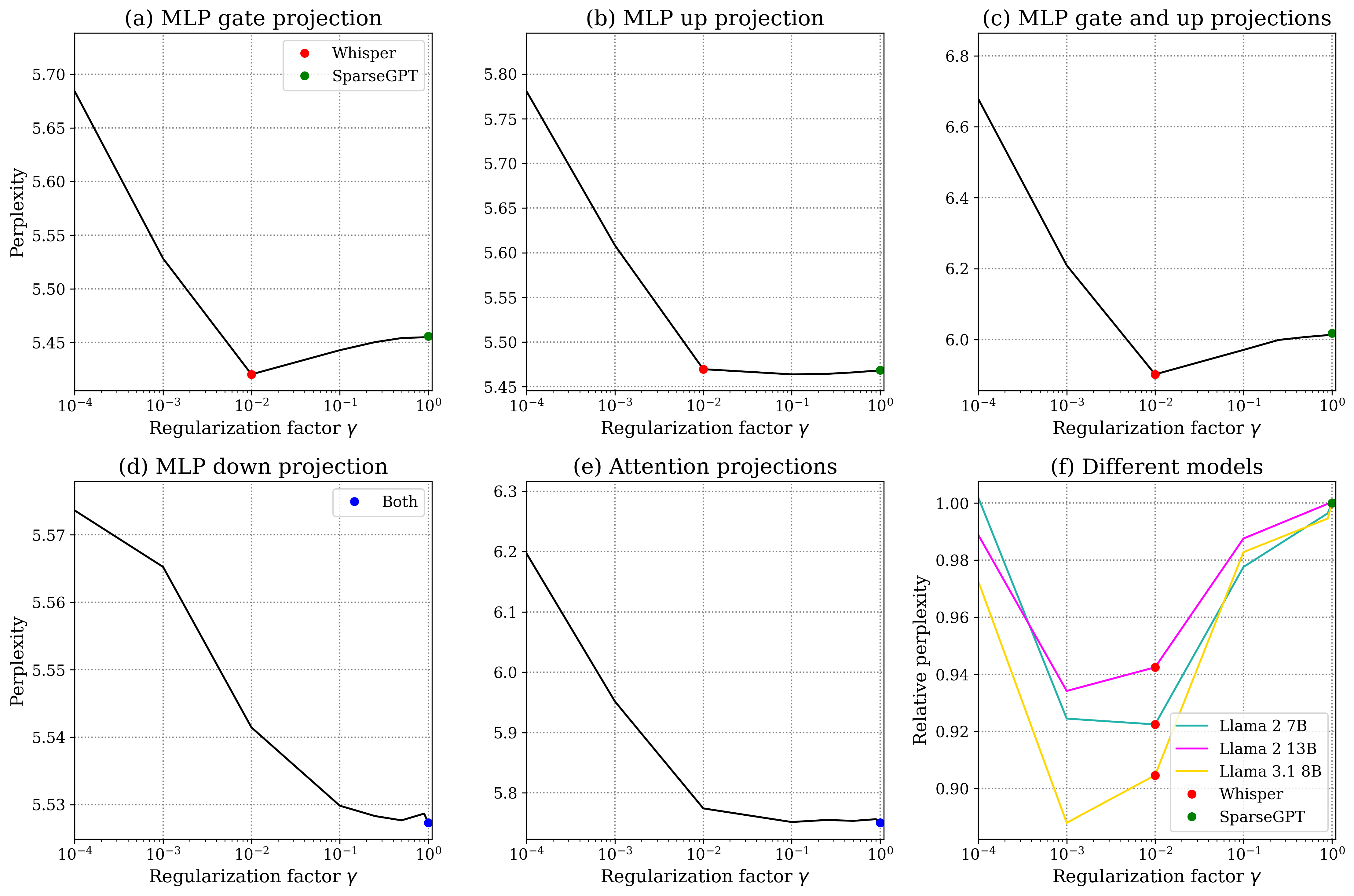}
    \caption{Effect of $\gamma$ for different transformer components. Perplexity when sparsifying only the MLP gate projection in every transformer block (a), only the MLP up projection in every transformer block (b), both the MLP gate and up projections in every transformer block (c), only the MLP down projection in every transformer block (d), and attention projections in every transformer block (e). (a-e) Analysis performed in Llama 2 7B. Perplexity for end to end sparsification of Llama 2 7B, Llama 2 13B, and Llama 3.1 8B with different $\gamma$ in \mymethodSO{} (f). The $\gamma$ used for all experiments in the main paper, 0.01, is marked, as well as the effective $\gamma$ for SparseGPT of 1. Layers and models pruned to 2:4 sparsity.}
    \label{fig:gamma_ablation}
\end{figure}

For the MLP gate and up projections, both separately and jointly, we observe a decrease in perplexity compared to $\gamma = 1$ (SparseGPT) as $\gamma$ decreases. This improvement in performance occurs up to a point, after which perplexity once again rises (\Cref{fig:gamma_ablation}a-c). This likely corresponds to the point in which the reconstruction of $\mathbf{W}\Delta\mathbf{X}$ comes at too high of a cost to the reconstruction of $\mathbf{W}\mathbf{X}$ itself, as observed in \Cref{fig:gamma_explanation}.

However, for the MLP down and attention projections, we observe that the overall perplexity for any $\gamma < 1.0$ is worse than $\gamma = 1$, which aligns with the observation that Wasserstein neurons themselves are concentrated in the up and gate projections \citep{sawmya2025wasserstein, kong2025negative}.

We then measure perplexity for different models when sparsifying them end to end to 2:4 sparsity for different $\gamma$ with \mymethodSO{}. We find that decreasing $\gamma$ improves performance, with perplexity decreasing consistently, up to a point. For $\gamma$ below $\approx0.001$, we see perplexity increase and performance degrade. We thus use $\gamma = 0.01$ for all \mymethodSO{} experiments in the main paper. Optimizing $\gamma$ per model or per layer remains interesting future work.

\clearpage

\subsection{Mapping difficulty definition}
\label{sec:mapping_diff}
Mapping difficulty is a metric proposed by prior work \citep{sawmya2025wasserstein} to measure the ability of a neuron to differentiate similar inputs. We use it only to select for Wasserstein neurons for ablation analyses and as a validation that \mymethodFON{} preserves this ability, but do not use it directly to design our pruning algorithms. We provide the formulation below:

Consider a linear layer $\mathbf{W}\in\mathbb{R}^{n\times m}$ with $n$ output neurons and $m$ input channels. Let $\mathbf{X}=[\mathbf{x}_1,\ldots,\mathbf{x}_s]\in\mathbb{R}^{m\times s}$
denote the calibration inputs to the layer, where each column $\mathbf{x}_\ell\in\mathbb{R}^m$ is one input vector.

Let $\pi$ be a random permutation of $\{1,\ldots,s\}$ and let $\mathbf{X}_{\pi}=[\mathbf{x}_{\pi(1)},\ldots,\mathbf{x}_{\pi(s)}]$ be the corresponding column-permuted calibration matrix. We define $\Delta\mathbf{X}=\mathbf{X}-\mathbf{X}_{\pi}$,
so that the $\ell$-th column is the sampled pairwise difference $\Delta\mathbf{X}_{:,\ell}=\mathbf{x}_{\ell}-\mathbf{x}_{\pi(\ell)}$. For neuron $i$, the output difference induced by this pair is $\Delta y_{i,\ell} = \mathbf{w}_i \Delta\mathbf{X}_{:,\ell}$.

Compute the $\ell_2$ norm for each of the $\ell$ pairwise difference of input vectors, 
$
\|\Delta\mathbf{X}_{:,\ell}\|_2
$. This is normalized by the maximum $N_x = \operatornamewithlimits{max}\limits_{\ell \in \{1,\dots,s\}} \|\Delta\mathbf{X}_{:,\ell}\|_2$ to bound the norms between 0 and 1.

For each neuron $i$, compute the $\ell_1$ norm for each pairwise difference in output scalars, $|\Delta y_{i,\ell}| = |\mathbf{w}_i \Delta\mathbf{X}_{:,\ell}|$. This is normalized by the median $N_{y,i} =  \operatornamewithlimits{median}\limits_{\ell \in \{1,\dots,s\}}|\Delta y_{i,\ell}|$ to account for differences in scales between neurons.

The mapping difficulty of each neuron is then defined as the average of the per-pair ratio of the normalized output differences to the normalized input differences:

\[
\text{MD}_i = \operatornamewithlimits{mean}\limits_{\ell \in \{1,\dots,s\}} (\frac{|\Delta y_{i,\ell}|}{N_{y,i}}) / (\frac{\|\Delta\mathbf{X}_{:,\ell}\|_2}{N_x})
\]

This measures the degree of separation in output space that a neuron must provide for incoming inputs.

\clearpage

\subsection{From \mymethodFOL{} to \mymethodFON{}}
\label{sec:wisp_to_wispplus}

\begin{figure}[h]
    \centering 
    \includegraphics[width=0.8\linewidth]{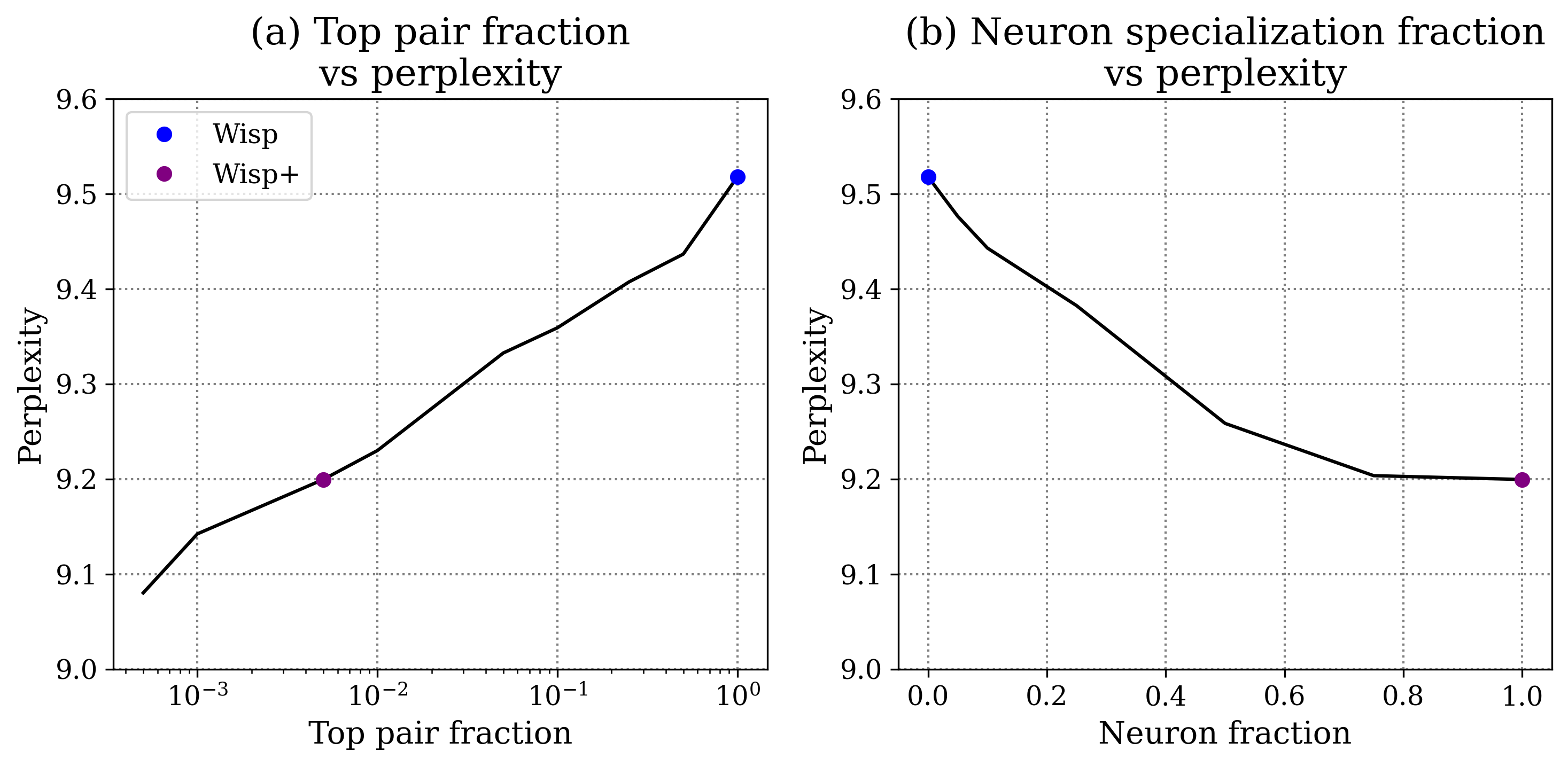}
    \caption{Effect of specialization degree on \mymethodFON{} performance. Decreasing the proportion of all pairs that are specialized upon increases performance (a). Increasing the proportion of neurons that are specialized increases performance (b). Analysis performed in Llama 2 7B at 2:4 sparsity.}
    \label{fig:fraction_ablations}
\end{figure}

For \mymethodFON{}, to better understand how the choice of $K$, the number of pairs each neuron is specialized on, impacts performance, we sweep $K$ and measure end to end perplexity in Llama 2 7B at 2:4 sparsity. For $K = s$, with $s$ being the total number of samples of pairwise differences, this essentially reduces to \mymethodFOL{}, with $\ell_1$ instead of $\ell_2$ norm. As the fraction of $s$ that $K$ comprises decreases, we observe a consistent decrease in perplexity, suggesting that increasing specialization yields increasing performance (\Cref{fig:fraction_ablations}a). We use $K = 0.005s$ in all experiments for \mymethodFON{} in the main body.

We also investigate how changing the number of neurons that are specialized within \mymethodFON{} impacts performance. We again measure end to end perplexity in Llama 2 7B at 2:4 sparsity, using $K = 0.005s$. For each gate and up projection, we sweep the proportion of neurons that we apply the \mymethodFON{} saliency to, and apply \mymethodFOL{} to the remainder. For a fraction of 1.0, this is equivalent to \mymethodFON{}, and for a fraction of 0.0, this is equivalent to \mymethodFOL{}.

To choose the neurons that we apply \mymethodFON{} to within each layer, we order them by their mapping difficulty. For all specialization fractions, we specialize the neurons with the highest mapping difficulty first to understand whether specialization most helps neurons that significantly separate similar inputs, or whether it helps in general. We follow the formulation of mapping difficulty established by \cite{sawmya2025wasserstein}, which we describe in \Cref{sec:mapping_diff}. We use this metric purely to understand the behavior of this algorithm under this ablation, and the metric does not factor into the pruning algorithms presented in the main body.

We observe that specializing neurons with higher mapping difficulty yields the largest marginal gains. Perplexity drops most steeply when moving from no specialization to specializing the neurons with the highest mapping difficulties. However, the improvement is not confined exclusively to this top subset. Perplexity continues to decrease as a larger fraction of neurons is specialized, suggesting that input differentiation is a broader property shared by many neurons in the gate and up projections, rather than a capability localized only to the Wasserstein neurons. At the same time, the curve flattens once most neurons have been specialized: the final gains from including the neurons with the lowest mapping difficulties, especially the bottom 20\%, are small. Thus, specialization is generally beneficial, but exhibits diminishing returns in the regime of least input separation. For consistency, we specialize all neurons in \mymethodFON{} in the main experiments. 

\clearpage

\subsection{Runtime}
\label{sec:runtime}

We report wall-clock pruning time for an entire model on an NVIDIA H100 GPU. The proposed methods do not introduce hidden computational blowups.

\begin{table}[h]
\centering
\begin{tabular}{lccccc}
\toprule
Model & Wanda & \mymethodFOL{} & \mymethodFON{} & SparseGPT & \mymethodSO{}\\
\midrule
Llama 2 7B & 48.7 s& 51.3 s& 144 s& 348 s& 374 s\\
Llama 2 13B & 83.3 s& 87.3 s& 258 s& 615 s& 670 s\\
Llama 2 70B & 420 s& 434 s& 1476 s& 3279 s& 3520 s\\
\midrule
\end{tabular}
\caption{Wall-clock time of each pruning method in seconds. Results collected over three trials.}
\label{tab:runtime}
\end{table}

\mymethodFOL{} has essentially the same runtime as Wanda, indicating that the layerwise difference statistic improves it while staying in the same practical cost regime. \mymethodFON{} is slower than \mymethodFOL{} because it selects neuron-specific pairs that are highly differentiated, but it remains substantially faster than SparseGPT. This makes \mymethodFON{} an intermediate point on the accuracy-cost frontier: it is more accurate than Wanda-style pruning while remaining cheaper than second-order reconstruction. Finally, \mymethodSO{} has nearly the same runtime as SparseGPT. Since \mymethodSO{} uses the same mask selection and weight update procedure as SparseGPT, this small overhead is from constructing $\mathbf{H}_\Delta$. The accuracy gains therefore come from changing the reconstruction geometry, not from a more expensive solver.

\clearpage

\subsection{Composability with RIA and ALPS}
\label{sec:composability_full}

To assess our difference-informed criteria in terms of their composability with and applicability to more recent pruning methods, we augment RIA \citep{zhang2024ria} and ALPS \citep{meng2024alps} to form RIA-\mymethodFOL{}, RIA-\mymethodFON{}, and ALPS-\mymethodSO{}.

RIA is a stronger first-order baseline than Wanda that uses the following saliency

\[
\mathbf{S}_{ij}^{\mathrm{RIA}}
=
(
\frac{|\mathbf{W}_{ij}|}{\|\mathbf{W}_{:j}\|_1}
+
\frac{|\mathbf{W}_{ij}|}{\|\mathbf{W}_{i:}\|_1}
)
\cdot
(\|\mathbf{X}_{j}\|_2)^{a}
\]

which normalizes weight importance across both rows and columns and scales activations. We augment RIA with the \mymethodFOL{} and \mymethodFON{} activation saliencies to yield RIA-\mymethodFOL{}, with criterion:
\[
(
\frac{|\mathbf{W}_{ij}|}{\|\mathbf{W}_{:j}\|_1}
+
\frac{|\mathbf{W}_{ij}|}{\|\mathbf{W}_{i:}\|_1}
) \cdot (\|\Delta \mathbf{X}_{j}\|_2)^a
\]
as well as RIA-\mymethodFON{}, with criterion:
\[
(
\frac{|\mathbf{W}_{ij}|}{\|\mathbf{W}_{:j}\|_1}
+
\frac{|\mathbf{W}_{ij}|}{\|\mathbf{W}_{i:}\|_1}
) \cdot 
(\frac{1}{|\mathcal{P}_i|}
\sum_{\ell\in\mathcal{P}_i}
|\Delta \mathbf{X}_{j,\ell}|)^a
\]
with $a=0.5$ for all three techniques, following the original RIA implementation. Like \mymethodFOL{} and \mymethodFON{}, these saliencies are applied only to the up and gate projections, leaving the original RIA saliency for the down and attention projections.

ALPS is a stronger reconstruction-based method than SparseGPT. For dense weights
$\mathbf{W}$ and sparse weights $\widehat{\mathbf{W}}$, it directly optimizes the
ridge-regularized layerwise reconstruction objective
\[
\min_{\widehat{\mathbf{W}}}
\frac{1}{2}
\|(\mathbf{W}-\widehat{\mathbf{W}})\mathbf{X}\|_F^2
+
\frac{\lambda_2}{2}
\|\mathbf{W}-\widehat{\mathbf{W}}\|_F^2
\qquad
\mathrm{s.t.}\quad
\|\widehat{\mathbf{W}}\|_0 \leq k
\]
The corresponding curvature matrix is
\[
\mathbf{H}_{\text{ALPS}}
=
\mathbf{X}\mathbf{X}^{T}+\lambda_2\mathbf{I}
\]
ALPS first uses the alternating direction method of multipliers (ADMM) to jointly
search for the sparse support and update the surviving weights, alternating between
a reconstruction update and projection onto the target sparsity pattern. Once the
support stabilizes, it fixes the mask and uses preconditioned conjugate gradient
(PCG) to further optimize the surviving weights.

To form ALPS-\mymethodSO{}, we retain this complete optimization procedure and all
original ALPS hyperparameters, but replace the standard activation curvature in the
MLP gate and up projections with our regularized difference-informed curvature, using the same $\gamma=0.01$ as \mymethodSO{}
\[
\widetilde{\mathbf{H}}
=
\gamma\mathbf{H}
+
(1-\gamma)
\frac{\mu}{\mu_{\Delta}}
\mathbf{H}_{\Delta},
\qquad
\mathbf{H}=\mathbf{X}\mathbf{X}^{T},
\qquad
\mathbf{H}_{\Delta}
=
\Delta\mathbf{X}\Delta\mathbf{X}^{T}
\]
Accordingly, ALPS-\mymethodSO{} uses
\[
\mathbf{H}_{\text{ALPS-\mymethodSO{}}}
=
\widetilde{\mathbf{H}}+\lambda_2\mathbf{I}
\]
throughout both the ADMM and PCG stages, while retaining the standard ALPS
curvature for all other projections. Thus, ALPS improves the optimization of the
sparse reconstruction problem, whereas \mymethodSO{} changes its geometry from
primarily preserving absolute layer outputs to primarily preserving their
differences.

We observe that the difference-informed variants consistently improve their corresponding RIA and ALPS formulations with little additional cost, demonstrating composability with both stronger first-order saliency methods and stronger second-order reconstruction optimizers. Full results are reported in \Cref{sec:composability_tables}.
\clearpage

\subsection{Pseudocode for \mymethodFOL{}, \mymethodFON{}, \mymethodSO{}}
\label{sec:pseudocode}

Consider a linear layer with weight matrix
$\mathbf{W}\in\mathbb{R}^{n\times m}$, where $n$ is the number of output neurons
and $m$ is the number of input channels. Let
$\mathbf{X}=[\mathbf{x}_1,\ldots,\mathbf{x}_s]\in\mathbb{R}^{m\times s}$
denote the calibration inputs to the layer, where $s$ is the number of calibration
samples and each column $\mathbf{x}_\ell\in\mathbb{R}^m$ is one input vector. Note that for \mymethodFOL{} and \mymethodFON{}, we follow the precedent established by Wanda and prune each neuron to the same sparsity level, as we do for Wanda itself \citep{sun2024wanda}.

All three routines are applied only to the MLP up and gate projections. For attention projections and MLP down projections, we retain the corresponding standard baseline method: Wanda for first-order update-free pruning and SparseGPT for second-order pruning with weight updates.

\begin{algorithm}
\caption{\mymethodFOL{}: Layerwise Difference-Informed Saliency}
\begin{algorithmic}[1]
    \Require
    \Statex Inputs: Calibration activations $\mathbf{X}=[\mathbf{x}_1,\ldots,\mathbf{x}_s]\in\mathbb{R}^{m\times s}$
    \Statex Weights: Layer weights $\mathbf{W}\in\mathbb{R}^{n\times m}$
    \Ensure Saliency matrix $\mathbf{S}^{\text{\mymethodFOL}}\in\mathbb{R}^{n\times m}$

    \State Sample a random permutation $\pi$ of $\{1,\ldots,s\}$
    \State Construct $\mathbf{X}_{\pi}=[\mathbf{x}_{\pi(1)},\ldots,\mathbf{x}_{\pi(s)}]$
    \State Compute sampled pairwise differences $\Delta\mathbf{X}\gets \mathbf{X}-\mathbf{X}_{\pi}$

    \For{$j=1,\ldots,m$}
        \State Compute channel difference statistic $d_j\gets \|\Delta\mathbf{X}_{j,:}\|_2$
    \EndFor

    \For{$i=1,\ldots,n$}
        \For{$j=1,\ldots,m$}
            \State $\mathbf{S}^{\text{\mymethodFOL}}_{ij}\gets |\mathbf{W}_{ij}|\cdot d_j$
        \EndFor
    \EndFor

    \State \Return $\mathbf{S}^{\text{\mymethodFOL}}$
\end{algorithmic}
\end{algorithm}

\begin{algorithm}
\caption{\mymethodFON{}: Neuronwise Difference-Informed Saliency}
\begin{algorithmic}[1]
    \Require
    \Statex Inputs: Calibration activations $\mathbf{X}=[\mathbf{x}_1,\ldots,\mathbf{x}_s]\in\mathbb{R}^{m\times s}$
    \Statex Weights: Layer weights $\mathbf{W}\in\mathbb{R}^{n\times m}$
    \Statex Hyperparameters: Number of selected pairs $K$, by default $K=0.005s$
    \Ensure Saliency matrix $\mathbf{S}^{\text{\mymethodFON}}\in\mathbb{R}^{n\times m}$

    \State Sample a random permutation $\pi$ of $\{1,\ldots,s\}$
    \State Construct $\mathbf{X}_{\pi}=[\mathbf{x}_{\pi(1)},\ldots,\mathbf{x}_{\pi(s)}]$
    \State Compute sampled pairwise differences $\Delta\mathbf{X}\gets \mathbf{X}-\mathbf{X}_{\pi}$

    \For{$\ell=1,\ldots,s$}
        \State Compute input-pair distance $d_{\ell}\gets \|\Delta\mathbf{X}_{:,\ell}\|_2$
    \EndFor

    \For{$i=1,\ldots,n$}
        \For{$\ell=1,\ldots,s$}
            \State Compute output separation $\Delta y_{i,\ell}\gets \mathbf{w}_i\Delta\mathbf{X}_{:,\ell}$
            \State Compute normalized separation $r_{i,\ell}\gets \dfrac{|\Delta y_{i,\ell}|}{d_{\ell}}$, ignore sample if $d_\ell = 0$
        \EndFor

        \State Select neuron-specific pairs
        \[
            \mathcal{P}_i\gets \operatorname{TopK}_{\ell\in\{1,\ldots,s\}} r_{i,\ell}
        \]

        \For{$j=1,\ldots,m$}
            \State Compute selected channel statistic
            \[
                a_{ij}\gets
                \frac{1}{|\mathcal{P}_i|}
                \sum_{\ell\in\mathcal{P}_i}
                |\Delta\mathbf{X}_{j,\ell}|
            \]
            \State $\mathbf{S}^{\text{\mymethodFON}}_{ij}\gets |\mathbf{W}_{ij}|\cdot a_{ij}$
        \EndFor
    \EndFor

    \State \Return $\mathbf{S}^{\text{\mymethodFON}}$
\end{algorithmic}
\end{algorithm}

\begin{algorithm}
\caption{\mymethodSO{}: Difference-Informed Hessian Reconstruction}
\begin{algorithmic}[1]
    \Require
    \Statex Inputs: Calibration activations $\mathbf{X}=[\mathbf{x}_1,\ldots,\mathbf{x}_s]\in\mathbb{R}^{m\times s}$
    \Statex Weights: Layer weights $\mathbf{W}\in\mathbb{R}^{n\times m}$
    \Statex Hyperparameters: Target sparsity $\rho$, regularization factor $\gamma$ (by default $\gamma = 0.01)$, damping $\lambda$
    \Ensure Pruned and reconstructed weights $\mathbf{W}'$

    \State Sample a random permutation $\pi$ of $\{1,\ldots,s\}$
    \State Construct $\mathbf{X}_{\pi}=[\mathbf{x}_{\pi(1)},\ldots,\mathbf{x}_{\pi(s)}]$
    \State Compute sampled pairwise differences $\Delta\mathbf{X}\gets \mathbf{X}-\mathbf{X}_{\pi}$

    \State Compute standard activation Hessian $\mathbf{H}\gets \mathbf{X}\mathbf{X}^{T}$
    \State Compute difference Hessian $\mathbf{H}_{\Delta}\gets \Delta\mathbf{X}\Delta\mathbf{X}^{T}$

    \State Compute average diagonal scales
    \[
        \mu\gets \frac{\operatorname{tr}(\mathbf{H})}{m},
        \qquad
        \mu_{\Delta}\gets \frac{\operatorname{tr}(\mathbf{H}_{\Delta})}{m}
    \]

    \State Form regularized difference-informed Hessian
    \[
        \widetilde{\mathbf{H}}
        \gets
        \gamma\mathbf{H}
        +
        (1-\gamma)
        \frac{\mu}{\mu_{\Delta}}
        \mathbf{H}_{\Delta}
    \]

    \State Compute damped inverse Hessian
    \[
        \mathbf{G}\gets
        \left(\widetilde{\mathbf{H}}+\lambda\mathbf{I}\right)^{-1}
    \]

    \For{$i=1,\ldots,n$}
        \For{$j=1,\ldots,m$}
            \State Compute pruning score
            \[
                \mathbf{S}^{\text{\mymethodSO}}_{ij}
                \gets
                \frac{|\mathbf{W}_{ij}|^2}{\mathbf{G}_{jj}}
            \]
        \EndFor
    \EndFor

    \State Select weights to prune according to $\mathbf{S}^{\text{\mymethodSO}}$ and target sparsity $\rho$
    \State Reconstruct remaining weights using the SparseGPT iterative mask and weight update rule with Hessian $\widetilde{\mathbf{H}}$
    \State Let $\mathbf{W}'$ denote the resulting sparse reconstructed weights

    \State \Return $\mathbf{W}'$
\end{algorithmic}
\end{algorithm}

\clearpage
\subsection{Raw perplexity and benchmark evaluation values}
\label{sec:raw_scores}

\subsubsection{Sparsity sweep for Llama 2}

\begin{table}[h]
\centering
\begin{tabular}{lcccccc}
\toprule
Method & $50\%$ & $55\%$ & $60\%$ & $65\%$ & $70\%$ & $75\%$\\
\midrule
\multicolumn{7}{c}{\underline{Llama 2 7B}} \\
Magnitude & $14.90$ & $59.26$ & $3.7e3$ & $3.7e4$ & $5.2e4$ & $4.8e4$ \\
Wanda & $6.31$ & $7.12$ & $9.21$ & $18.69$ & $60.12$ & $287.6$ \\
\mymethodFOL & $\underline{6.25}$ & $6.93$ & $8.44$ & $14.55$ & $43.56$ & $228.05$ \\
\mymethodFON & $6.26$ & $\underline{6.89}$ & $\underline{8.27}$ & $\underline{13.19}$ & $\underline{36.00}$ & $\underline{169.09}$ \\
SparseGPT & $6.09$ & $6.67$ & $7.86$ & $10.60$ & $15.89$ & $26.70$ \\
\mymethodSO & $\mathbf{6.05}$ & $\mathbf{6.50}$ & $\mathbf{7.35}$ & $\mathbf{9.39}$ & $\mathbf{13.63}$ & $\mathbf{23.06}$ \\

\midrule
\multicolumn{7}{c}{\underline{Llama 2 13B}} \\
Magnitude & $6.37$ & $7.83$ & $11.22$ & $26.25$ & $275.2$ & $7.8e3$ \\
Wanda & $5.47$ & $6.02$ & $7.27$ & $11.40$ & $33.48$ & $82.05$ \\
\mymethodFOL & $\underline{5.42}$ & $5.88$ & $6.86$ & $9.58$ & $21.94$ & $72.72$ \\
\mymethodFON & $\underline{5.42}$ & $\underline{5.86}$ & $\underline{6.76}$ & $\underline{9.06}$ & $\underline{18.93}$ & $\underline{56.28}$ \\
SparseGPT & $5.33$ & $5.76$ & $6.59$ & $8.43$ & $11.96$ & $20.36$ \\
\mymethodSO & $\mathbf{5.29}$ & $\mathbf{5.62}$ & $\mathbf{6.24}$ & $\mathbf{7.49}$ & $\mathbf{10.44}$ & $\mathbf{17.17}$ \\

\midrule
\multicolumn{7}{c}{\underline{Llama 2 70B}} \\
Magnitude & $4.99$ & $6.14$ & $8.20$ & $13.73$ & $1.5e3$ & $2.5e4$ \\
Wanda & $3.91$ & $4.27$ & $4.80$ & $5.88$ & $8.75$ & $17.95$ \\
\mymethodFOL{} & $\underline{3.89}$ & $4.22$ & $4.70$ & $5.60$ & $7.91$ & $14.84$ \\
\mymethodFON & $\underline{3.89}$ & $\underline{4.20}$ & $\underline{4.66}$ & $\underline{5.44}$ & $\underline{7.39}$ & $\underline{13.04}$ \\
SparseGPT & $3.81$ & $4.10$ & $4.52$ & $5.22$ & $6.72$ & $9.48$ \\
\mymethodSO & $\mathbf{\underline{3.77}}$ & $\mathbf{\underline{4.01}}$ & $\mathbf{\underline{4.35}}$ & $\mathbf{\underline{4.81}}$ & $\mathbf{\underline{5.74}}$ & $\mathbf{\underline{8.10}}$ \\
\midrule
\end{tabular}
\caption{Language modeling performance on Llama 2 family for different unstructured sparsities. Bold indicates best performance and underline indicates best performance without weight updates. Raw values for \Cref{fig:llama_2_curve}.}
\label{tab:sparsity_sweep}
\end{table}

\clearpage

\subsubsection{Perplexity and evaluations for Llama 2}

\begin{table}[h]
\centering
\resizebox{\linewidth}{!}{
\begin{tabular}{lcccccccc}
\toprule
Method & Perplexity & Mean & ARC Challenge & MathQA & HellaSwag & MMLU & TruthfulQA & WinoGrande \\
\midrule
Dense & $5.12$ & $53.22$ & $53.16$ & $28.31$ & $78.61$ & $46.58$ & $38.96$ & $73.72$ \\
\midrule
& \multicolumn{8}{c}{\underline{2:4 Sparsity}}\\

Magnitude & $54.39$ & $38.16$ & $30.29$ & $23.08$ & $47.24$ & $27.05$ & $\mathbf{\underline{42.23}}$ & $59.04$ \\
Wanda & $10.53 \pm 0.02$ & $39.39 \pm 0.02$ & $34.19 \pm 0.27$ & $23.31 \pm 0.03$ & $53.27 \pm 0.11$ & $27.91 \pm 0.23$ & $39.89 \pm 0.12$ & $57.80 \pm 0.42$ \\
\mymethodFOL{} & $9.52 \pm 0.01$ & $40.41 \pm 0.05$ & $35.78 \pm 0.08$ & $23.81 \pm 0.17$ & $56.58 \pm 0.14$ & $\underline{28.88 \pm 0.07}$ & $38.63 \pm 0.20$ & $58.77 \pm 0.39$ \\
\mymethodFON{} & $\underline{9.20 \pm 0.00}$ & $\underline{40.79 \pm 0.07}$ & $\underline{36.72 \pm 0.20}$ & $\underline{23.93 \pm 0.14}$ & $\underline{57.81 \pm 0.08}$ & $27.80 \pm 0.38$ & $38.58 \pm 0.23$ & $\underline{59.88 \pm 0.21}$ \\
SparseGPT & $8.18 \pm 0.01$ & $41.19 \pm 0.05$ & $37.71 \pm 0.55$ & $\mathbf{24.77 \pm 0.23}$ & $55.26 \pm 0.10$ & $28.74 \pm 0.41$ & $37.45 \pm 0.12$ & $63.22 \pm 0.72$ \\
\mymethodSO{} & $\mathbf{7.56 \pm 0.01}$ & $\mathbf{41.98 \pm 0.10}$ & $\mathbf{38.59 \pm 0.54}$ & $24.13 \pm 0.25$ & $\mathbf{58.18 \pm 0.03}$ & $\mathbf{29.44 \pm 0.35}$ & $37.06 \pm 0.18$ & $\mathbf{64.46 \pm 0.52}$ \\
\midrule

& \multicolumn{8}{c}{\underline{50\% Sparsity}}\\
Magnitude & $14.89$ & $45.25$ & $42.83$ & $24.05$ & $67.01$ & $29.32$ & $\mathbf{\underline{41.19}}$ & $67.09$ \\
Wanda & $6.31 \pm 0.00$ & $47.41 \pm 0.08$ & $47.47 \pm 0.30$ & $26.00 \pm 0.07$ & $72.01 \pm 0.04$ & $\underline{32.40 \pm 0.44}$ & $37.27 \pm 0.07$ & $\underline{69.32 \pm 0.14}$ \\
\mymethodFOL{} & $\underline{6.25 \pm 0.00}$ & $47.65 \pm 0.11$ & $48.01 \pm 0.25$ & $26.47 \pm 0.02$ & $72.76 \pm 0.02$ & $31.94 \pm 0.34$ & $38.03 \pm 0.09$ & $68.69 \pm 0.21$ \\
\mymethodFON{} & $\underline{6.25 \pm 0.00}$ & $\underline{47.91 \pm 0.02}$ & $\mathbf{\underline{48.49 \pm 0.16}}$ & $\underline{26.68 \pm 0.10}$ & $\mathbf{\underline{73.42 \pm 0.05}}$ & $32.02 \pm 0.09$ & $38.08 \pm 0.19$ & $68.75 \pm 0.33$ \\
SparseGPT & $6.09 \pm 0.00$ & $48.35 \pm 0.08$ & $47.04 \pm 0.53$ & $\mathbf{26.73 \pm 0.22}$ & $71.72 \pm 0.03$ & $\mathbf{36.39 \pm 0.20}$ & $37.47 \pm 0.49$ & $\mathbf{70.72 \pm 0.16}$ \\
\mymethodSO{} & $\mathbf{6.04 \pm 0.00}$ & $\mathbf{48.62 \pm 0.16}$ & $46.96 \pm 0.46$ & $26.71 \pm 0.23$ & $73.06 \pm 0.08$ & $35.89 \pm 0.35$ & $38.59 \pm 0.41$ & $70.53 \pm 0.44$ \\
\midrule

& \multicolumn{8}{c}{\underline{65\% Sparsity}}\\
Magnitude & $36925.58$ & $32.76$ & $26.96$ & $19.13$ & $25.83$ & $24.20$ & $\mathbf{\underline{51.09}}$ & $49.33$ \\
Wanda & $18.81 \pm 0.08$ & $33.92 \pm 0.08$ & $24.26 \pm 0.37$ & $23.63 \pm 0.09$ & $35.28 \pm 0.11$ & $\mathbf{\underline{27.51 \pm 0.04}}$ & $40.36 \pm 0.07$ & $52.46 \pm 0.15$ \\
\mymethodFOL{} & $14.61 \pm 0.04$ & $35.19 \pm 0.14$ & $27.02 \pm 0.03$ & $23.90 \pm 0.18$ & $41.44 \pm 0.25$ & $27.08 \pm 0.16$ & $38.19 \pm 0.22$ & $53.54 \pm 0.32$ \\
\mymethodFON{} & $\underline{13.17 \pm 0.02}$ & $\underline{36.05 \pm 0.07}$ & $\underline{28.24 \pm 0.31}$ & $\underline{23.91 \pm 0.08}$ & $\underline{44.74 \pm 0.07}$ & $27.48 \pm 0.08$ & $37.49 \pm 0.08$ & $\underline{54.46 \pm 0.32}$ \\
SparseGPT & $10.64 \pm 0.04$ & $38.36 \pm 0.10$ & $31.88 \pm 0.17$ & $24.09 \pm 0.00$ & $47.64 \pm 0.11$ & $27.48 \pm 0.30$ & $38.61 \pm 0.34$ & $60.46 \pm 0.36$ \\
\mymethodSO{} & $\mathbf{9.46 \pm 0.02}$ & $\mathbf{39.67 \pm 0.21}$ & $\mathbf{34.76 \pm 0.38}$ & $\mathbf{24.10 \pm 0.34}$ & $\mathbf{52.16 \pm 0.19}$ & $27.21 \pm 0.50$ & $38.65 \pm 0.75$ & $\mathbf{61.17 \pm 0.24}$ \\

\midrule
\end{tabular}
}
\caption{Language modeling and evaluation performance for Llama 2 7B. Bold indicates best performance and underline indicates best performance without weight updates. Results acquired over three trials. Error indicates one standard error of the mean. Raw values for \Cref{tab:llama_main}.}
\label{tab:llama_2_7B_full}
\end{table}

\clearpage

\begin{table}[h]
\centering
\resizebox{\linewidth}{!}{
\begin{tabular}{lcccccccc}
\toprule
Method & Perplexity & Mean & ARC Challenge & MathQA & HellaSwag & MMLU & TruthfulQA & WinoGrande \\
\midrule
Dense & $4.57$ & $57.01$ & $59.64$ & $31.83$ & $82.19$ & $55.35$ & $36.90$ & $76.16$ \\
\midrule
& \multicolumn{8}{c}{\underline{2:4 Sparsity}}\\

Magnitude & $8.32$ & $43.93$ & $39.93$ & $23.72$ & $\mathbf{\underline{67.37}}$ & $29.65$ & $\mathbf{\underline{38.35}}$ & $64.56$ \\
Wanda & $7.79 \pm 0.01$ & $43.94 \pm 0.04$ & $39.73 \pm 0.20$ & $25.84 \pm 0.16$ & $62.14 \pm 0.14$ & $34.70 \pm 0.16$ & $37.13 \pm 0.06$ & $64.09 \pm 0.25$ \\
\mymethodFOL{} & $7.36 \pm 0.01$ & $45.49 \pm 0.05$ & $41.70 \pm 0.25$ & $\underline{26.23 \pm 0.20}$ & $65.01 \pm 0.05$ & $35.36 \pm 0.14$ & $37.49 \pm 0.10$ & $67.14 \pm 0.16$ \\
\mymethodFON{} & $\underline{7.24 \pm 0.01}$ & $\underline{46.00 \pm 0.05}$ & $\underline{42.52 \pm 0.19}$ & $26.05 \pm 0.19$ & $66.11 \pm 0.12$ & $\underline{35.85 \pm 0.29}$ & $37.53 \pm 0.14$ & $\underline{67.96 \pm 0.41}$ \\
SparseGPT & $6.69 \pm 0.02$ & $44.93 \pm 0.26$ & $40.53 \pm 0.10$ & $26.07 \pm 0.30$ & $61.65 \pm 0.31$ & $36.56 \pm 0.68$ & $36.80 \pm 0.22$ & $67.96 \pm 0.23$ \\
\mymethodSO{} & $\mathbf{6.29 \pm 0.01}$ & $\mathbf{46.60 \pm 0.16}$ & $\mathbf{43.03 \pm 1.16}$ & $\mathbf{26.38 \pm 0.29}$ & $65.41 \pm 0.09$ & $\mathbf{37.41 \pm 0.33}$ & $38.07 \pm 0.46$ & $\mathbf{69.32 \pm 0.30}$ \\
\midrule

& \multicolumn{8}{c}{\underline{50\% Sparsity}}\\
Magnitude & $6.37$ & $51.12$ & $49.49$ & $27.07$ & $75.81$ & $43.96$ & $\mathbf{\underline{39.45}}$ & $70.96$ \\
Wanda & $5.47 \pm 0.00$ & $52.87 \pm 0.04$ & $53.33 \pm 0.09$ & $29.32 \pm 0.13$ & $76.83 \pm 0.03$ & $47.59 \pm 0.12$ & $36.24 \pm 0.06$ & $73.90 \pm 0.25$ \\
\mymethodFOL{} & $\underline{5.42 \pm 0.00}$ & $\mathbf{\underline{53.28 \pm 0.04}}$ & $\mathbf{\underline{54.12 \pm 0.10}}$ & $\mathbf{\underline{29.40 \pm 0.11}}$ & $77.53 \pm 0.03$ & $\mathbf{\underline{47.73 \pm 0.09}}$ & $36.76 \pm 0.09$ & $74.16 \pm 0.09$ \\
\mymethodFON{} & $\underline{5.42 \pm 0.00}$ & $53.21 \pm 0.02$ & $\mathbf{\underline{54.12 \pm 0.08}}$ & $29.32 \pm 0.10$ & $\mathbf{\underline{78.03 \pm 0.09}}$ & $47.62 \pm 0.10$ & $35.99 \pm 0.17$ & $\mathbf{\underline{74.19 \pm 0.20}}$ \\
SparseGPT & $5.34 \pm 0.00$ & $52.84 \pm 0.09$ & $52.99 \pm 0.05$ & $28.78 \pm 0.10$ & $76.38 \pm 0.07$ & $46.05 \pm 0.23$ & $39.03 \pm 0.20$ & $73.85 \pm 0.31$ \\
\mymethodSO{} & $\mathbf{5.28 \pm 0.00}$ & $53.14 \pm 0.10$ & $53.44 \pm 0.15$ & $29.11 \pm 0.05$ & $77.48 \pm 0.07$ & $46.42 \pm 0.13$ & $39.14 \pm 0.48$ & $73.27 \pm 0.30$ \\
\midrule

& \multicolumn{8}{c}{\underline{65\% Sparsity}}\\
Magnitude & $26.24$ & $35.87$ & $27.99$ & $22.21$ & $43.20$ & $25.07$ & $\mathbf{\underline{43.46}}$ & $53.28$ \\
Wanda & $11.33 \pm 0.06$ & $36.72 \pm 0.09$ & $29.69 \pm 0.21$ & $25.20 \pm 0.28$ & $45.42 \pm 0.17$ & $27.63 \pm 0.04$ & $37.19 \pm 0.15$ & $55.17 \pm 0.30$ \\
\mymethodFOL{} & $9.59 \pm 0.01$ & $38.90 \pm 0.13$ & $32.91 \pm 0.45$ & $25.43 \pm 0.05$ & $51.67 \pm 0.01$ & $27.85 \pm 0.22$ & $37.14 \pm 0.15$ & $58.43 \pm 0.22$ \\
\mymethodFON{} & $\underline{9.04 \pm 0.01}$ & $\underline{39.71 \pm 0.09}$ & $\underline{34.67 \pm 0.25}$ & $\mathbf{\underline{25.44 \pm 0.05}}$ & $\underline{53.77 \pm 0.09}$ & $\underline{27.94 \pm 0.37}$ & $36.76 \pm 0.24$ & $\underline{59.69 \pm 0.25}$ \\
SparseGPT & $8.41 \pm 0.01$ & $41.11 \pm 0.09$ & $37.49 \pm 0.22$ & $24.84 \pm 0.28$ & $54.61 \pm 0.28$ & $29.99 \pm 0.53$ & $37.73 \pm 0.11$ & $62.01 \pm 0.23$ \\
\mymethodSO{} & $\mathbf{7.54 \pm 0.02}$ & $\mathbf{42.57 \pm 0.13}$ & $\mathbf{38.91 \pm 0.71}$ & $25.25 \pm 0.19$ & $\mathbf{58.64 \pm 0.18}$ & $\mathbf{30.76 \pm 0.65}$ & $37.74 \pm 0.33$ & $\mathbf{64.09 \pm 0.23}$ \\

\midrule
\end{tabular}
}
\caption{Language modeling and evaluation performance for Llama 2 13B. Bold indicates best performance and underline indicates best performance without weight updates. Results acquired over three trials. Error indicates one standard error of the mean. Raw values for \Cref{tab:llama_main}.}
\label{tab:llama_2_13B_full}
\end{table}

\begin{table}[h]
\centering
\resizebox{\linewidth}{!}{
\begin{tabular}{lcccccccc}
\toprule
Method & Perplexity & Mean & ARC Challenge & MathQA & HellaSwag & MMLU & TruthfulQA & WinoGrande \\
\midrule
Dense & $3.12$ & $65.17$ & $67.32$ & $38.32$ & $87.27$ & $69.65$ & $44.82$ & $83.66$ \\
\midrule
& \multicolumn{8}{c}{\underline{2:4 Sparsity}}\\

Magnitude & $6.33$ & $54.83$ & $55.97$ & $30.35$ & $78.36$ & $50.73$ & $\mathbf{\underline{39.90}}$ & $73.64$ \\
Wanda & $4.97 \pm 0.00$ & $57.13 \pm 0.04$ & $\underline{59.50 \pm 0.06}$ & $31.89 \pm 0.17$ & $78.63 \pm 0.01$ & $55.96 \pm 0.09$ & $38.73 \pm 0.15$ & $78.08 \pm 0.07$ \\
\mymethodFOL{} & $4.86 \pm 0.00$ & $57.57 \pm 0.04$ & $59.16 \pm 0.15$ & $31.98 \pm 0.07$ & $79.41 \pm 0.05$ & $56.92 \pm 0.16$ & $39.63 \pm 0.11$ & $78.32 \pm 0.16$ \\
\mymethodFON{} & $\underline{4.81 \pm 0.00}$ & $\underline{57.79 \pm 0.04}$ & $59.04 \pm 0.05$ & $\underline{32.19 \pm 0.02}$ & $\underline{79.72 \pm 0.11}$ & $\underline{57.45 \pm 0.17}$ & $39.77 \pm 0.21$ & $\underline{78.56 \pm 0.23}$ \\
SparseGPT & $4.71 \pm 0.01$ & $57.05 \pm 0.04$ & $58.96 \pm 0.05$ & $\mathbf{32.51 \pm 0.32}$ & $78.20 \pm 0.14$ & $57.26 \pm 0.31$ & $37.43 \pm 0.38$ & $77.95 \pm 0.14$ \\
\mymethodSO{} & $\mathbf{4.51 \pm 0.00}$ & $\mathbf{58.39 \pm 0.03}$ & $\mathbf{60.69 \pm 0.44}$ & $32.45 \pm 0.25$ & $\mathbf{79.91 \pm 0.13}$ & $\mathbf{58.82 \pm 0.23}$ & $39.65 \pm 0.47$ & $\mathbf{78.80 \pm 0.27}$ \\
\midrule

& \multicolumn{8}{c}{\underline{50\% Sparsity}}\\
Magnitude & $4.99$ & $60.54$ & $62.54$ & $34.04$ & $83.47$ & $62.24$ & $42.11$ & $78.85$ \\
Wanda & $3.91 \pm 0.00$ & $62.52 \pm 0.03$ & $\mathbf{\underline{66.38 \pm 0.17}}$ & $34.66 \pm 0.09$ & $84.90 \pm 0.03$ & $64.35 \pm 0.12$ & $42.83 \pm 0.07$ & $81.98 \pm 0.14$ \\
\mymethodFOL{} & $3.89 \pm 0.00$ & $\underline{62.81 \pm 0.02}$ & $66.15 \pm 0.16$ & $34.80 \pm 0.04$ & $85.17 \pm 0.06$ & $\underline{64.93 \pm 0.05}$ & $\underline{43.45 \pm 0.07}$ & $82.37 \pm 0.30$ \\
\mymethodFON{} & $\underline{3.88 \pm 0.00}$ & $62.76 \pm 0.05$ & $65.96 \pm 0.18$ & $\underline{34.85 \pm 0.08}$ & $\underline{85.23 \pm 0.02}$ & $64.76 \pm 0.09$ & $43.33 \pm 0.14$ & $\mathbf{\underline{82.45 \pm 0.05}}$ \\
SparseGPT & $3.81 \pm 0.01$ & $62.66 \pm 0.15$ & $66.33 \pm 0.33$ & $34.96 \pm 0.13$ & $84.93 \pm 0.07$ & $64.81 \pm 0.16$ & $42.76 \pm 0.53$ & $82.19 \pm 0.19$ \\
\mymethodSO{} & $\mathbf{3.77 \pm 0.00}$ & $\mathbf{62.96 \pm 0.08}$ & $66.27 \pm 0.49$ & $\mathbf{35.15 \pm 0.36}$ & $\mathbf{85.24 \pm 0.02}$ & $\mathbf{65.12 \pm 0.08}$ & $\mathbf{43.53 \pm 0.41}$ & $\mathbf{82.45 \pm 0.23}$ \\
\midrule

& \multicolumn{8}{c}{\underline{65\% Sparsity}}\\
Magnitude & $13.73$ & $47.02$ & $48.55$ & $26.87$ & $70.63$ & $31.45$ & $\underline{39.99}$ & $64.64$ \\
Wanda & $5.89 \pm 0.01$ & $54.31 \pm 0.12$ & $55.75 \pm 0.12$ & $31.64 \pm 0.01$ & $73.45 \pm 0.08$ & $51.37 \pm 0.24$ & $36.40 \pm 0.09$ & $77.27 \pm 0.18$ \\
\mymethodFOL{} & $5.60 \pm 0.00$ & $55.64 \pm 0.01$ & $56.68 \pm 0.17$ & $31.60 \pm 0.11$ & $75.41 \pm 0.01$ & $\underline{53.07 \pm 0.11}$ & $38.49 \pm 0.23$ & $78.58 \pm 0.16$ \\
\mymethodFON{} & $\underline{5.44 \pm 0.00}$ & $\underline{55.83 \pm 0.00}$ & $\underline{57.14 \pm 0.25}$ & $\underline{31.71 \pm 0.12}$ & $\underline{75.90 \pm 0.06}$ & $52.96 \pm 0.05$ & $38.54 \pm 0.14$ & $\underline{78.74 \pm 0.17}$ \\
SparseGPT & $5.22 \pm 0.01$ & $55.76 \pm 0.08$ & $57.25 \pm 0.36$ & $\mathbf{31.76 \pm 0.25}$ & $75.71 \pm 0.10$ & $53.53 \pm 0.12$ & $38.56 \pm 0.42$ & $77.77 \pm 0.25$ \\
\mymethodSO{} & $\mathbf{4.83 \pm 0.01}$ & $\mathbf{57.71 \pm 0.02}$ & $\mathbf{60.18 \pm 0.10}$ & $31.75 \pm 0.16$ & $\mathbf{78.72 \pm 0.06}$ & $\mathbf{55.68 \pm 0.23}$ & $\mathbf{40.56 \pm 0.49}$ & $\mathbf{79.37 \pm 0.16}$ \\

\midrule
\end{tabular}
}
\caption{Language modeling and evaluation performance for Llama 2 70B. Bold indicates best performance and underline indicates best performance without weight updates. Results acquired over three trials. Error indicates one standard error of the mean. Raw values for \Cref{tab:llama_main,tab:llama_2_70B}.}
\label{tab:llama_2_70B_full}
\end{table}

\clearpage

\subsubsection{Perplexity and evaluations for Llama 3.1}

\begin{table}[h]
\centering
\resizebox{\linewidth}{!}{
\begin{tabular}{lcccccccc}
\toprule
Method & Perplexity & Mean & ARC Challenge & MathQA & HellaSwag & MMLU & TruthfulQA & WinoGrande \\
\midrule
Dense & $5.84$ & $61.52$ & $57.68$ & $39.56$ & $81.71$ & $66.29$ & $45.17$ & $78.69$ \\
\midrule
& \multicolumn{8}{c}{\underline{2:4 Sparsity}}\\
Magnitude & $765.32$ & $32.77$ & $22.78$ & $\underline{25.09}$ & $29.85$ & $25.44$ & $\mathbf{\underline{42.77}}$ & $50.67$ \\
Wanda & $18.87 \pm 0.02$ & $36.81 \pm 0.15$ & $27.96 \pm 0.08$ & $24.85 \pm 0.38$ & $44.87 \pm 0.19$ & $29.47 \pm 0.23$ & $38.08 \pm 0.12$ & $55.67 \pm 0.51$ \\
\mymethodFOL{} & $17.65 \pm 0.14$ & $38.03 \pm 0.12$ & $30.06 \pm 0.08$ & $25.07 \pm 0.10$ & $47.50 \pm 0.13$ & $\underline{30.81 \pm 0.31}$ & $38.15 \pm 0.30$ & $56.62 \pm 0.21$ \\
\mymethodFON{} & $\underline{16.70 \pm 0.06}$ & $\underline{38.40 \pm 0.10}$ & $\underline{31.09 \pm 0.40}$ & $24.98 \pm 0.06$ & $\underline{48.49 \pm 0.03}$ & $30.38 \pm 0.44$ & $37.87 \pm 0.10$ & $\underline{57.62 \pm 0.47}$ \\
SparseGPT & $11.19 \pm 0.08$ & $41.62 \pm 0.24$ & $35.44 \pm 0.20$ & $\mathbf{25.99 \pm 0.33}$ & $52.61 \pm 0.46$ & $34.53 \pm 0.99$ & $37.83 \pm 0.37$ & $\mathbf{63.33 \pm 0.07}$ \\
\mymethodSO{} & $\mathbf{10.18 \pm 0.04}$ & $\mathbf{42.24 \pm 0.26}$ & $\mathbf{36.01 \pm 0.86}$ & $25.72 \pm 0.44$ & $\mathbf{55.63 \pm 0.26}$ & $\mathbf{34.78 \pm 0.26}$ & $38.01 \pm 0.58$ & $63.30 \pm 0.12$ \\
\midrule

& \multicolumn{8}{c}{\underline{50\% Sparsity}}\\
Magnitude & $150.89$ & $38.66$ & $32.85$ & $25.86$ & $41.03$ & $31.98$ & $\mathbf{\underline{47.67}}$ & $52.57$ \\
Wanda & $8.76 \pm 0.01$ & $50.79 \pm 0.04$ & $\underline{46.08 \pm 0.26}$ & $\underline{31.76 \pm 0.14}$ & $68.36 \pm 0.08$ & $49.68 \pm 0.14$ & $38.16 \pm 0.18$ & $70.72 \pm 0.12$ \\
\mymethodFOL{} & $8.62 \pm 0.01$ & $\underline{51.31 \pm 0.05}$ & $45.93 \pm 0.36$ & $31.55 \pm 0.16$ & $69.70 \pm 0.04$ & $\underline{51.43 \pm 0.10}$ & $38.46 \pm 0.27$ & $70.77 \pm 0.11$ \\
\mymethodFON{} & $\underline{8.50 \pm 0.01}$ & $51.25 \pm 0.07$ & $45.79 \pm 0.27$ & $31.39 \pm 0.18$ & $\underline{70.01 \pm 0.11}$ & $50.73 \pm 0.31$ & $38.44 \pm 0.33$ & $\underline{71.11 \pm 0.08}$ \\
SparseGPT & $7.80 \pm 0.01$ & $53.16 \pm 0.21$ & $49.43 \pm 0.53$ & $\mathbf{32.40 \pm 0.28}$ & $72.56 \pm 0.10$ & $54.45 \pm 0.22$ & $38.73 \pm 0.60$ & $\mathbf{71.40 \pm 0.42}$ \\
\mymethodSO{} & $\mathbf{7.64 \pm 0.00}$ & $\mathbf{53.49 \pm 0.03}$ & $\mathbf{50.40 \pm 0.16}$ & $32.27 \pm 0.38$ & $\mathbf{73.95 \pm 0.13}$ & $\mathbf{54.52 \pm 0.34}$ & $39.19 \pm 0.11$ & $70.64 \pm 0.21$ \\
\midrule

& \multicolumn{8}{c}{\underline{65\% Sparsity}}\\
Magnitude & $33929.21$ & $\underline{33.19}$ & $\underline{25.60}$ & $20.87$ & $25.91$ & $25.03$ & $\mathbf{\underline{49.34}}$ & $\underline{52.41}$ \\
Wanda & $36.48 \pm 0.33$ & $32.44 \pm 0.03$ & $21.08 \pm 0.34$ & $23.14 \pm 0.06$ & $31.95 \pm 0.05$ & $26.77 \pm 0.22$ & $41.74 \pm 0.06$ & $49.99 \pm 0.27$ \\
\mymethodFOL{} & $34.90 \pm 0.42$ & $32.89 \pm 0.02$ & $21.62 \pm 0.16$ & $\underline{24.21 \pm 0.04}$ & $32.46 \pm 0.21$ & $27.00 \pm 0.29$ & $41.84 \pm 0.11$ & $50.20 \pm 0.14$ \\
\mymethodFON{} & $\underline{30.09 \pm 0.29}$ & $33.10 \pm 0.07$ & $22.50 \pm 0.12$ & $23.86 \pm 0.12$ & $\underline{33.50 \pm 0.22}$ & $\underline{27.02 \pm 0.16}$ & $41.43 \pm 0.08$ & $50.28 \pm 0.45$ \\
SparseGPT & $14.51 \pm 0.09$ & $38.95 \pm 0.13$ & $30.57 \pm 0.47$ & $\mathbf{25.96 \pm 0.40}$ & $47.06 \pm 0.02$ & $\mathbf{29.71 \pm 0.16}$ & $40.91 \pm 0.39$ & $\mathbf{59.46 \pm 0.62}$ \\
\mymethodSO{} & $\mathbf{13.07 \pm 0.07}$ & $\mathbf{39.37 \pm 0.28}$ & $\mathbf{32.59 \pm 0.78}$ & $25.33 \pm 0.06$ & $\mathbf{50.05 \pm 0.02}$ & $28.98 \pm 1.04$ & $40.21 \pm 0.32$ & $59.06 \pm 0.13$ \\

\midrule
\end{tabular}
}
\caption{Language modeling and evaluation performance for Llama 3.1 8B. Bold indicates best performance and underline indicates best performance without weight updates. Results acquired over three trials. Error indicates one standard error of the mean. Raw values for \Cref{tab:llama_main}.}
\label{tab:llama_31_8B_full}
\end{table}

\begin{table}[h]
\centering
\resizebox{\linewidth}{!}{
\begin{tabular}{lcccccccc}
\toprule
Method & Perplexity & Mean & ARC Challenge & MathQA & HellaSwag & MMLU & TruthfulQA & WinoGrande \\
\midrule
Dense & $2.64$ & $70.62$ & $69.88$ & $51.76$ & $88.03$ & $78.86$ & $49.74$ & $85.48$ \\
\midrule
& \multicolumn{8}{c}{\underline{2:4 Sparsity}}\\
Magnitude & $1089.38$ & $39.34$ & $24.57$ & $21.98$ & $45.90$ & $33.82$ & $\mathbf{\underline{42.40}}$ & $67.40$ \\
Wanda & $7.84 \pm 0.02$ & $56.76 \pm 0.05$ & $57.31 \pm 0.12$ & $35.11 \pm 0.07$ & $72.79 \pm 0.23$ & $59.53 \pm 0.57$ & $40.35 \pm 0.43$ & $75.48 \pm 0.25$ \\
\mymethodFOL{} & $7.61 \pm 0.02$ & $57.82 \pm 0.10$ & $58.19 \pm 0.15$ & $35.11 \pm 0.32$ & $75.76 \pm 0.22$ & $60.22 \pm 0.27$ & $42.06 \pm 0.42$ & $75.56 \pm 0.42$ \\
\mymethodFON{} & $\underline{7.51 \pm 0.01}$ & $\underline{58.01 \pm 0.17}$ & $\underline{58.22 \pm 0.21}$ & $\underline{35.50 \pm 0.10}$ & $\mathbf{\underline{76.01 \pm 0.25}}$ & $\underline{60.66 \pm 0.24}$ & $41.83 \pm 0.73$ & $\underline{75.82 \pm 0.41}$ \\
SparseGPT & $6.78 \pm 0.01$ & $56.98 \pm 0.39$ & $56.26 \pm 0.60$ & $36.17 \pm 0.26$ & $71.90 \pm 0.92$ & $61.24 \pm 0.40$ & $40.04 \pm 0.44$ & $76.30 \pm 0.50$ \\
\mymethodSO{} & $\mathbf{6.40 \pm 0.02}$ & $\mathbf{58.22 \pm 0.61}$ & $\mathbf{58.76 \pm 0.83}$ & $\mathbf{36.53 \pm 0.68}$ & $74.18 \pm 1.59$ & $\mathbf{62.59 \pm 0.19}$ & $40.39 \pm 0.69$ & $\mathbf{76.85 \pm 0.37}$ \\
\midrule

& \multicolumn{8}{c}{\underline{50\% Sparsity}}\\
Magnitude & $17.91$ & $54.14$ & $57.34$ & $27.40$ & $66.98$ & $66.14$ & $41.26$ & $65.75$ \\
Wanda & $5.29 \pm 0.00$ & $\underline{64.97 \pm 0.07}$ & $64.93 \pm 0.30$ & $\mathbf{\underline{44.81 \pm 0.01}}$ & $\underline{81.83 \pm 0.16}$ & $73.18 \pm 0.11$ & $\underline{42.87 \pm 0.25}$ & $\underline{82.19 \pm 0.14}$ \\
\mymethodFOL{} & $5.25 \pm 0.00$ & $64.92 \pm 0.06$ & $65.67 \pm 0.21$ & $44.33 \pm 0.11$ & $81.13 \pm 0.17$ & $\mathbf{\underline{73.60 \pm 0.04}}$ & $42.75 \pm 0.21$ & $82.03 \pm 0.27$ \\
\mymethodFON{} & $\underline{5.23 \pm 0.00}$ & $64.95 \pm 0.08$ & $\underline{65.96 \pm 0.09}$ & $44.77 \pm 0.37$ & $81.25 \pm 0.12$ & $73.47 \pm 0.01$ & $42.83 \pm 0.32$ & $81.43 \pm 0.09$ \\
SparseGPT & $4.94 \pm 0.00$ & $65.77 \pm 0.05$ & $65.76 \pm 0.30$ & $44.48 \pm 0.26$ & $83.56 \pm 0.27$ & $73.07 \pm 0.05$ & $44.57 \pm 0.13$ & $\mathbf{83.19 \pm 0.12}$ \\
\mymethodSO{} & $\mathbf{4.93 \pm 0.01}$ & $\mathbf{65.97 \pm 0.10}$ & $\mathbf{66.18 \pm 0.41}$ & $44.78 \pm 0.12$ & $\mathbf{84.37 \pm 0.24}$ & $73.44 \pm 0.04$ & $\mathbf{44.58 \pm 0.68}$ & $82.48 \pm 0.09$ \\
\midrule

& \multicolumn{8}{c}{\underline{65\% Sparsity}}\\
Magnitude & $8331.90$ & $32.60$ & $26.88$ & $19.73$ & $26.68$ & $23.19$ & $\mathbf{\underline{48.58}}$ & $50.51$ \\
Wanda & $9.83 \pm 0.10$ & $49.41 \pm 0.14$ & $49.06 \pm 0.49$ & $32.47 \pm 0.18$ & $58.43 \pm 0.35$ & $47.76 \pm 0.36$ & $39.77 \pm 0.21$ & $68.98 \pm 0.46$ \\
\mymethodFOL{} & $9.25 \pm 0.03$ & $51.17 \pm 0.32$ & $51.59 \pm 0.12$ & $32.47 \pm 0.14$ & $61.91 \pm 0.50$ & $\underline{50.15 \pm 0.59}$ & $41.57 \pm 0.42$ & $69.32 \pm 0.37$ \\
\mymethodFON{} & $\underline{9.01 \pm 0.06}$ & $\underline{51.55 \pm 0.21}$ & $\underline{51.91 \pm 0.70}$ & $\underline{33.04 \pm 0.22}$ & $\underline{62.16 \pm 0.77}$ & $50.05 \pm 0.37$ & $41.72 \pm 0.16$ & $\underline{70.43 \pm 0.35}$ \\
SparseGPT & $7.29 \pm 0.03$ & $55.34 \pm 0.20$ & $\mathbf{54.72 \pm 0.86}$ & $\mathbf{34.09 \pm 0.20}$ & $70.57 \pm 0.64$ & $56.51 \pm 0.63$ & $39.86 \pm 0.84$ & $76.27 \pm 0.41$ \\
\mymethodSO{} & $\mathbf{6.88 \pm 0.01}$ & $\mathbf{56.00 \pm 0.17}$ & $54.27 \pm 0.13$ & $33.61 \pm 0.20$ & $\mathbf{73.62 \pm 0.22}$ & $\mathbf{58.57 \pm 0.72}$ & $39.25 \pm 0.42$ & $\mathbf{76.66 \pm 0.37}$ \\

\midrule
\end{tabular}
}
\caption{Language modeling and evaluation performance for Llama 3.1 70B. Bold indicates best performance and underline indicates best performance without weight updates. Results acquired over three trials. Error indicates one standard error of the mean. Raw values for \Cref{tab:llama_main}.}
\end{table}

\begin{table}[h]
\centering
\begin{tabular}{lc}
\toprule
Method & Perplexity\\
\midrule
Dense & $1.34$\\
\midrule
& \multicolumn{1}{c}{\underline{2:4 Sparsity}}\\
Magnitude & $10.55$ \\
Wanda & $4.60 \pm 0.00$ \\
\mymethodFOL{} & $4.53 \pm 0.00$ \\
\mymethodFON{} & $\underline{4.50 \pm 0.00}$\\
SparseGPT & $4.20 \pm 0.01$ \\
\mymethodSO{} & $\mathbf{4.06 \pm 0.01}$ \\
\midrule

& \multicolumn{1}{c}{\underline{50\% Sparsity}}\\
Magnitude & $40.27$ \\
Wanda & $2.75 \pm 0.00$ \\
\mymethodFOL{} & $2.73 \pm 0.00$ \\
\mymethodFON{} & $\underline{2.72 \pm 0.00}$ \\
SparseGPT & $2.59 \pm 0.00$ \\
\mymethodSO{} & $\mathbf{2.57 \pm 0.00}$ \\
\midrule

& \multicolumn{1}{c}{\underline{65\% Sparsity}}\\
Magnitude & $11214.91$ \\
Wanda & $5.64 \pm 0.00$ \\
\mymethodFOL{} & $5.51 \pm 0.01$ \\
\mymethodFON{} & $\underline{5.48 \pm 0.01}$ \\
SparseGPT & $4.86 \pm 0.00$ \\
\mymethodSO{} & $\mathbf{4.68 \pm 0.01}$ \\

\midrule
\end{tabular}
\caption{Language modeling performance for Llama 3.1 405B. We report only perplexity due to the prohibitive cost of downstream evaluations at this scale. Bold indicates best performance and underline indicates best performance without weight updates. Results acquired over three trials. Error indicates one standard error of the mean. Raw values for \Cref{tab:llama_main}.}
\end{table}

\clearpage

\subsubsection{Perplexity and evaluations for Mistral 7B v0.3 and Qwen3}

\begin{table}[h]
\centering
\resizebox{\linewidth}{!}{
\begin{tabular}{lcccccccc}
\toprule
Method & PPL & Mean & ARC-C & MathQA & HS & MMLU & TQA & WinoG \\
\midrule
Dense & $4.95$ & $60.56$ & $60.41$ & $35.51$ & $83.05$ & $63.58$ & $42.61$ & $78.22$ \\
Magnitude & $13.49$ & $42.12$ & $38.05$ & $25.33$ & $\underline{60.10}$ & $28.87$ & $\underline{38.56}$ & $61.80$ \\
Wanda & $9.64 \pm 0.04$ & $41.49 \pm 0.10$ & $37.12 \pm 0.18$ & $27.20 \pm 0.12$ & $55.71 \pm 0.16$ & $31.24 \pm 0.35$ & $37.08 \pm 0.04$ & $60.56 \pm 0.39$ \\
\mymethodFOL{} & $8.97 \pm 0.03$ & $42.74 \pm 0.07$ & $38.20 \pm 0.17$ & $\mathbf{\underline{27.40 \pm 0.15}}$ & $58.08 \pm 0.04$ & $32.62 \pm 0.26$ & $38.14 \pm 0.10$ & $\underline{61.98 \pm 0.23}$ \\
\mymethodFON{} & $\underline{8.64 \pm 0.03}$ & $\underline{43.05 \pm 0.04}$ & $\underline{38.94 \pm 0.15}$ & $26.97 \pm 0.11$ & $58.79 \pm 0.04$ & $\underline{33.74 \pm 0.41}$ & $37.92 \pm 0.12$ & $61.96 \pm 0.30$ \\
SparseGPT & $7.47 \pm 0.03$ & $44.79 \pm 0.39$ & $41.33 \pm 0.23$ & $27.26 \pm 0.32$ & $59.19 \pm 0.17$ & $38.25 \pm 1.23$ & $38.56 \pm 0.37$ & $64.14 \pm 0.33$ \\
\mymethodSO{} & $\mathbf{6.91 \pm 0.03}$ & $\mathbf{45.97 \pm 0.21}$ & $\mathbf{43.23 \pm 0.58}$ & $26.71 \pm 0.10$ & $\mathbf{62.27 \pm 0.15}$ & $\mathbf{39.73 \pm 0.71}$ & $\mathbf{39.29 \pm 0.18}$ & $\mathbf{64.61 \pm 0.14}$ \\
\midrule
\end{tabular}
}
\caption{Language modeling and evaluation performance for Mistral 7B v0.3 at 2:4 sparsity. Bold indicates best performance and underline indicates best performance without weight updates. Results acquired over three trials. Error indicates one standard error of the mean. Raw values for \Cref{tab:mistral_qwen}.}
\label{tab:mistral_full}
\end{table}

\begin{table}[h]
\centering
\resizebox{\linewidth}{!}{
\begin{tabular}{lcccccccc}
\toprule
Method & PPL & Mean & ARC-C & MathQA & HS & MMLU & TQA & WinoG \\
\midrule
Dense & $6.51$ & $68.25$ & $68.00$ & $54.34$ & $79.63$ & $78.79$ & $52.33$ & $76.40$ \\
Magnitude & $754.43$ & $31.92$ & $21.08$ & $19.63$ & $27.10$ & $24.46$ & $\mathbf{\underline{48.11}}$ & $51.14$ \\
Wanda & $11.33 \pm 0.02$ & $50.31 \pm 0.08$ & $47.38 \pm 0.40$ & $\mathbf{\underline{39.21 \pm 0.17}}$ & $56.16 \pm 0.12$ & $56.80 \pm 0.20$ & $40.83 \pm 0.25$ & $\underline{61.48 \pm 0.16}$ \\
\mymethodFOL{} & $11.18 \pm 0.00$ & $\underline{50.49 \pm 0.10}$ & $\underline{47.64 \pm 0.27}$ & $38.96 \pm 0.08$ & $56.43 \pm 0.05$ & $\mathbf{\underline{57.75 \pm 0.12}}$ & $40.85 \pm 0.17$ & $61.30 \pm 0.50$ \\
\mymethodFON{} & $\underline{10.86 \pm 0.01}$ & $50.46 \pm 0.08$ & $47.24 \pm 0.06$ & $38.62 \pm 0.14$ & $\underline{56.80 \pm 0.11}$ & $57.40 \pm 0.16$ & $41.32 \pm 0.12$ & $61.40 \pm 0.48$ \\
SparseGPT & $9.04 \pm 0.02$ & $51.12 \pm 0.18$ & $49.37 \pm 0.53$ & $35.68 \pm 0.45$ & $59.23 \pm 0.15$ & $54.50 \pm 0.85$ & $42.25 \pm 0.28$ & $\mathbf{65.67 \pm 0.25}$ \\
\mymethodSO{} & $\mathbf{8.82 \pm 0.01}$ & $\mathbf{51.45 \pm 0.12}$ & $\mathbf{49.72 \pm 0.36}$ & $35.81 \pm 0.13$ & $\mathbf{60.23 \pm 0.10}$ & $55.08 \pm 0.35$ & $42.72 \pm 0.53$ & $65.11 \pm 0.16$ \\
\midrule
\end{tabular}
}
\caption{Language modeling and evaluation performance for Qwen3 8B Base at 2:4 sparsity. Bold indicates best performance and underline indicates best performance without weight updates. Results acquired over three trials. Error indicates one standard error of the mean. Raw values for \Cref{tab:mistral_qwen}.}
\label{tab:qwen3_8B_full}
\end{table}

\begin{table}[h]
\centering
\resizebox{\linewidth}{!}{
\begin{tabular}{lcccccccc}
\toprule
Method & PPL & Mean & ARC-C & MathQA & HS & MMLU & TQA & WinoG \\
\midrule
Dense & $5.95$ & $71.70$ & $69.20$ & $61.78$ & $82.50$ & $82.23$ & $55.07$ & $79.40$ \\
Magnitude & $121.28$ & $41.57$ & $39.08$ & $24.86$ & $40.91$ & $45.66$ & $\mathbf{\underline{45.48}}$ & $53.43$ \\
Wanda & $8.81 \pm 0.00$ & $56.66 \pm 0.14$ & $55.38 \pm 0.23$ & $42.12 \pm 0.67$ & $64.98 \pm 0.05$ & $65.55 \pm 0.33$ & $42.53 \pm 0.09$ & $69.40 \pm 0.21$ \\
\mymethodFOL{} & $8.68 \pm 0.00$ & $56.65 \pm 0.03$ & $54.92 \pm 0.17$ & $42.42 \pm 0.21$ & $65.45 \pm 0.14$ & $66.00 \pm 0.27$ & $41.94 \pm 0.15$ & $69.19 \pm 0.22$ \\
\mymethodFON{} & $\underline{8.55 \pm 0.00}$ & $\underline{57.03 \pm 0.16}$ & $\underline{55.75 \pm 0.11}$ & $\mathbf{\underline{42.94 \pm 0.65}}$ & $\underline{65.57 \pm 0.10}$ & $\mathbf{\underline{66.42 \pm 0.23}}$ & $41.99 \pm 0.12$ & $\underline{69.51 \pm 0.09}$ \\
SparseGPT & $7.80 \pm 0.01$ & $57.82 \pm 0.08$ & $57.17 \pm 0.87$ & $40.70 \pm 0.32$ & $67.03 \pm 0.08$ & $66.11 \pm 0.09$ & $44.13 \pm 0.10$ & $71.80 \pm 0.30$ \\
\mymethodSO{} & $\mathbf{7.65 \pm 0.01}$ & $\mathbf{58.11 \pm 0.12}$ & $\mathbf{57.25 \pm 0.65}$ & $40.08 \pm 0.39$ & $\mathbf{68.34 \pm 0.05}$ & $66.37 \pm 0.17$ & $44.58 \pm 0.41$ & $\mathbf{72.03 \pm 0.47}$ \\
\midrule
\end{tabular}
}
\caption{Language modeling and evaluation performance for Qwen3 14B Base at 2:4 sparsity. Bold indicates best performance and underline indicates best performance without weight updates. Results acquired over three trials. Error indicates one standard error of the mean. Raw values for \Cref{tab:mistral_qwen}.}
\label{tab:qwen3_14B_full}
\end{table}

\clearpage

\subsubsection{Perplexity for Granite 4.1}
\label{sec:granite_ppl}

\begin{table}[h]
\centering
\begin{tabular}{lcccc}
\toprule
& \multicolumn{4}{c}{\underline{Sparsity}}\\
Method & 0\% & 2:4 & 50\% & 65\%\\
\midrule
Dense & $6.53$ &  &  &  \\
Magnitude &  & $297.40$ & $244.80$ & $1620449.25$ \\
Wanda &  & $12.06 \pm 0.01$ & $8.59 \pm 0.00$ & $16.60 \pm 0.02$ \\
\mymethodFOL{} &  & $11.69 \pm 0.01$ & $8.58 \pm 0.01$ & $15.68 \pm 0.03$ \\
\mymethodFON{} &  & $\underline{11.52 \pm 0.02}$ & $\underline{8.56 \pm 0.01}$ & $\underline{14.98 \pm 0.02}$ \\
SparseGPT &  & $9.80 \pm 0.00$ & $8.19 \pm 0.01$ & $12.84 \pm 0.07$ \\
\mymethodSO{} &  & $\mathbf{9.56 \pm 0.01}$ & $\mathbf{8.09 \pm 0.00}$ & $\mathbf{11.27 \pm 0.06}$ \\
\midrule
\end{tabular}
\caption{Language modeling performance for Granite 4.1 8B Base. Bold indicates best performance and underline indicates best performance without weight updates. Results acquired over three trials. Error indicates one standard error of the mean.}
\label{tab:granite_41_8b_full}
\end{table}

\begin{table}[h]
\centering
\begin{tabular}{lcccc}
\toprule
& \multicolumn{4}{c}{\underline{Sparsity}}\\
Method & 0\% & 2:4 & 50\% & 65\%\\
\midrule
Dense & $5.29$ &  &  &  \\
Magnitude &  & $28.98$ & $40.08$ & $140585.61$ \\
Wanda &  & $8.41 \pm 0.00$ & $\underline{6.69 \pm 0.00}$ & $9.89 \pm 0.01$ \\
\mymethodFOL{} &  & $8.36 \pm 0.00$ & $\underline{6.69 \pm 0.01}$ & $9.87 \pm 0.02$ \\
\mymethodFON{} &  & $\underline{8.35 \pm 0.00}$ & $6.70 \pm 0.01$ & $\underline{9.65 \pm 0.03}$ \\
SparseGPT &  & $7.56 \pm 0.01$ & $6.46 \pm 0.01$ & $8.44 \pm 0.01$ \\
\mymethodSO{} &  & $\mathbf{7.49 \pm 0.01}$ & $\mathbf{6.45 \pm 0.00}$ & $\mathbf{8.17 \pm 0.01}$ \\
\midrule
\end{tabular}
\caption{Language modeling performance for Granite 4.1 30B Base. Bold indicates best performance and underline indicates best performance without weight updates. Results acquired over three trials. Error indicates one standard error of the mean.}
\label{tab:granite_41_30b_full}
\end{table}

\clearpage

\subsubsection{Perplexity for Llama 2 with stronger pruning formulations}
\label{sec:composability_tables}
\begin{table}[h]
\centering
\begin{tabular}{lcccc}
\toprule
& \multicolumn{3}{c}{\underline{Sparsity}} & \\
Method & 2:4 & 50\% & 65\% & Time (s)\\
\midrule
RIA & 10.24 $\pm$ 0.01 & 6.25 $\pm$ 0.00 & 18.23 $\pm$ 0.03 & 53.0 \\
RIA-Wisp & 9.57 $\pm$ 0.01 & 6.22 $\pm$ 0.00 & 15.83 $\pm$ 0.01 & 56.0 \\
RIA-Wisp+ & \underline{9.34 $\pm$ 0.01} & \underline{6.21 $\pm$ 0.00} & \underline{14.52 $\pm$ 0.04} & 149 \\
ALPS & 7.22 $\pm$ 0.02 & \textbf{5.88 $\pm$ 0.01} & 8.24 $\pm$ 0.03 & 1059 \\
ALPS-Whisper & \textbf{7.04 $\pm$ 0.01} & 5.89 $\pm$ 0.01 & \textbf{7.94 $\pm$ 0.01} & 1017 \\
\midrule
\end{tabular}
\caption{Language modeling performance for Llama 2 7B with stronger pruning methods. Bold indicates best performance and underline indicates best performance without weight updates. Results acquired over three trials. Error indicates one standard error of the mean. Raw values for \Cref{tab:composability}.}
\label{tab:llama_2_7b_ppl_compose}
\end{table}

\begin{table}[h]
\centering
\begin{tabular}{lcccc}
\toprule
& \multicolumn{3}{c}{\underline{Sparsity}} & \\
Method & 2:4 & 50\% & 65\% & Time (s)\\
\midrule
RIA & 7.48 $\pm$ 0.01 & 5.39 $\pm$ 0.00 & 10.22 $\pm$ 0.02 & 94.3 \\
RIA-Wisp & 7.19 $\pm$ 0.01 & 5.38 $\pm$ 0.00 & 9.30 $\pm$ 0.01 & 94.3 \\
RIA-Wisp+ & \underline{7.13 $\pm$ 0.01} & \underline{5.37 $\pm$ 0.00} & \underline{8.94 $\pm$ 0.00} & 265 \\
ALPS & 6.11 $\pm$ 0.01 & 5.17 $\pm$ 0.00 & 6.80 $\pm$ 0.01 & 2302 \\
ALPS-Whisper & \textbf{5.97 $\pm$ 0.01} & \textbf{5.15 $\pm$ 0.01} & \textbf{6.54 $\pm$ 0.01} & 2232 \\
\midrule
\end{tabular}
\caption{Language modeling performance for Llama 2 13B with stronger pruning methods. Bold indicates best performance and underline indicates best performance without weight updates. Results acquired over three trials. Error indicates one standard error of the mean. Raw values for \Cref{tab:composability}.}
\label{tab:llama_2_13b_ppl_compose}
\end{table}

\begin{table}[h]
\centering
\begin{tabular}{lcccc}
\toprule
& \multicolumn{3}{c}{\underline{Sparsity}} & \\
Method & 2:4 & 50\% & 65\% & Time (s)\\
\midrule
RIA & 4.91 $\pm$ 0.00 & 3.84 $\pm$ 0.00 & 5.79 $\pm$ 0.00 & 472 \\
RIA-Wisp & 4.85 $\pm$ 0.00 & \underline{3.83 $\pm$ 0.00} & 5.59 $\pm$ 0.00 & 462 \\
RIA-Wisp+ & \underline{4.81 $\pm$ 0.00} & \underline{3.83 $\pm$ 0.00} & \underline{5.46 $\pm$ 0.01} & 1516 \\
ALPS & 4.50 $\pm$ 0.01 & 3.74 $\pm$ 0.00 & 4.80 $\pm$ 0.01 & 17407 \\
ALPS-Whisper & \textbf{4.41 $\pm$ 0.01} & \textbf{3.72 $\pm$ 0.00} & \textbf{4.66 $\pm$ 0.01} & 17623 \\
\midrule
\end{tabular}
\caption{Language modeling performance for Llama 2 70B with stronger pruning methods. Bold indicates best performance and underline indicates best performance without weight updates. Results acquired over three trials. Error indicates one standard error of the mean. Raw values for \Cref{tab:composability}.}
\label{tab:llama_2_70b_ppl_compose}
\end{table}


\clearpage
\section*{NeurIPS Paper Checklist}

\begin{enumerate}

\item {\bf Claims}
    \item[] Question: Do the main claims made in the abstract and introduction accurately reflect the paper's contributions and scope?
    \item[] Answer: \answerYes{}
    \item[] Justification: The claims made in the abstract and introduction are supported by the methodology derivations and empirical results across different LLM families.

\item {\bf Limitations}
    \item[] Question: Does the paper discuss the limitations of the work performed by the authors?
    \item[] Answer: \answerYes{}
    \item[] Justification: The paper discusses limitations in \Cref{sec:limitations}.

\item {\bf Theory assumptions and proofs}
    \item[] Question: For each theoretical result, does the paper provide the full set of assumptions and a complete (and correct) proof?
    \item[] Answer: \answerNA{}
    \item[] Justification: The paper does not present formal theoretical results such as theorems or lemmas. It includes algebraic derivations motivating the proposed objective and Hessian construction, but these are methodological derivations rather than theoretical results requiring separately stated assumptions and proofs.

    \item {\bf Experimental result reproducibility}
    \item[] Question: Does the paper fully disclose all the information needed to reproduce the main experimental results of the paper to the extent that it affects the main claims and/or conclusions of the paper (regardless of whether the code and data are provided or not)?
    \item[] Answer: \answerYes{}
    \item[] Justification: The paper describes in detail all algorithms, including the selected hyperparameters, implemented in the main text, as well as provides the pseudocode in \Cref{sec:pseudocode}.

\item {\bf Open access to data and code}
    \item[] Question: Does the paper provide open access to the data and code, with sufficient instructions to faithfully reproduce the main experimental results, as described in supplemental material?
    \item[] Answer: \answerYes{}
    \item[] Justification: The complete codebase, including reproduction instructions, is provided at \texttt{\href{https://github.com/Shavit-Lab/Whisper}{https://github.com/Shavit-Lab/Whisper}}.

\item {\bf Experimental setting/details}
    \item[] Question: Does the paper specify all the training and test details (e.g., data splits, hyperparameters, how they were chosen, type of optimizer) necessary to understand the results?
    \item[] Answer: \answerYes{}
    \item[] Justification: All relevant calibration data, hyperparameter values, and task details are explained in the paper.

\item {\bf Experiment statistical significance}
    \item[] Question: Does the paper report error bars suitably and correctly defined or other appropriate information about the statistical significance of the experiments?
    \item[] Answer: \answerYes{}
    \item[] Justification: Error bars are defined in the full empirical result tables found in the appendix in \Cref{sec:raw_scores}.

\item {\bf Experiments compute resources}
    \item[] Question: For each experiment, does the paper provide sufficient information on the computer resources (type of compute workers, memory, time of execution) needed to reproduce the experiments?
    \item[] Answer: \answerYes{}
    \item[] Justification: The paper includes the GPU specifications used, as well as the wall clock times for various algorithms to run in \Cref{tab:runtime}.
    
\item {\bf Code of ethics}
    \item[] Question: Does the research conducted in the paper conform, in every respect, with the NeurIPS Code of Ethics \url{https://neurips.cc/public/EthicsGuidelines}?
    \item[] Answer: \answerYes{}
    \item[] Justification: The paper follows the NeurIPS Code of Ethics.

\item {\bf Broader impacts}
    \item[] Question: Does the paper discuss both potential positive societal impacts and negative societal impacts of the work performed?
    \item[] Answer: \answerNo{}
    \item[] Justification: The paper focuses on one-shot pruning and sparsification methods and does not include a dedicated broader impacts discussion. The work is methodological and not tied to a specific deployment setting. Its most direct potential positive impact is improved efficiency and accessibility of model inference, while potential negative impacts are indirect and would arise primarily through enabling more efficient deployment of existing models rather than introducing new model capabilities or application specific risks.
    
\item {\bf Safeguards}
    \item[] Question: Does the paper describe safeguards that have been put in place for responsible release of data or models that have a high risk for misuse (e.g., pre-trained language models, image generators, or scraped datasets)?
    \item[] Answer: \answerNA{}
    \item[] Justification: The paper does not release new pretrained models, generative systems, or scraped datasets that pose a high risk of misuse. The released artifact is an algorithmic pruning method and eventual codebase, so safeguards for controlled release of high-risk models or datasets are not applicable.

\item {\bf Licenses for existing assets}
    \item[] Question: Are the creators or original owners of assets (e.g., code, data, models), used in the paper, properly credited and are the license and terms of use explicitly mentioned and properly respected?
    \item[] Answer: \answerYes{}
    \item[] Justification: All prior methods and data used are explicitly cited.

\item {\bf New assets}
    \item[] Question: Are new assets introduced in the paper well documented and is the documentation provided alongside the assets?
    \item[] Answer: \answerNA{}
    \item[] Justification: The paper does not introduce new datasets, pretrained models, or benchmark assets. It proposes an algorithmic method for improving sparsification of existing pretrained LLMs, so asset documentation requirements are not applicable.

\item {\bf Crowdsourcing and research with human subjects}
    \item[] Question: For crowdsourcing experiments and research with human subjects, does the paper include the full text of instructions given to participants and screenshots, if applicable, as well as details about compensation (if any)? 
    \item[] Answer: \answerNA{}
    \item[] Justification: The paper does not involve any crowdsourcing nor research with human subjects.

\item {\bf Institutional review board (IRB) approvals or equivalent for research with human subjects}
    \item[] Question: Does the paper describe potential risks incurred by study participants, whether such risks were disclosed to the subjects, and whether Institutional Review Board (IRB) approvals (or an equivalent approval/review based on the requirements of your country or institution) were obtained?
    \item[] Answer: \answerNA{}
    \item[] Justification: The paper does not involve any crowdsourcing nor research with human subjects.

\item {\bf Declaration of LLM usage}
    \item[] Question: Does the paper describe the usage of LLMs if it is an important, original, or non-standard component of the core methods in this research? Note that if the LLM is used only for writing, editing, or formatting purposes and does \emph{not} impact the core methodology, scientific rigor, or originality of the research, declaration is not required.
    \item[] Answer: \answerNA{}
    \item[] Justification: The paper's contributions are entirely original. LLM usage is limited to editing and formatting purposes.

\end{enumerate}

\end{document}